\PassOptionsToPackage{table,dvipsnames}{xcolor}
\documentclass[sigconf]{acmart}
\usepackage{amsmath}
\usepackage{bbding}
\usepackage{utfsym}
\usepackage{pdfpages}
\usepackage{adjustbox}
\usepackage{multirow}
\usepackage[algo2e,linesnumbered,lined,boxed,commentsnumbered,ruled]{algorithm2e} 
\usepackage{url}
\usepackage{tabularx}
\usepackage{subcaption}
\usepackage{graphicx}
\usepackage{placeins}
\usepackage{float}
\usepackage{array}
\usepackage{stfloats}

\AtBeginDocument{%
  \providecommand\BibTeX{{%
    \normalfont B\kern-0.5em{\scshape i\kern-0.25em b}\kern-0.8em\TeX}}}

\copyrightyear{2025}
\acmYear{2025}
\setcopyright{rightsretained}
\acmConference[KDD '25] {Proceedings of the 31st ACM SIGKDD Conference on Knowledge Discovery and Data Mining V.2}{August 3--7, 2025}{Toronto, ON, Canada.}
\acmBooktitle{Proceedings of the 31st ACM SIGKDD Conference on Knowledge Discovery and Data Mining V.2 (KDD '25), August 3--7, 2025, Toronto, ON, Canada}
\acmISBN{978-1-4503-XXXX-X/18/06}
\acmDOI{10.1145/00000000.000000}

\begin{document}

\title[ErrorEraser: Unlearning Data Bias for Improved Continual Learning]%
{ErrorEraser: Unlearning Data Bias for Improved \\Continual Learning}




\author{Xuemei Cao}
\orcid{0000-0001-9083-8998}
\affiliation{%
  \institution{School of Computing and Artificial Intelligence, Southwestern University of Finance and Economics}
  \city{Chengdu}
  \country{China}
  \postcode{611130}
}
\email{caoxuemei.pqz@gmail.com}

\author{Hanlin Gu}
\authornote{Co-Corresponding Author.}
\orcid{0000-0001-8266-4561}
\affiliation{%
  \institution{WeBank}
  \city{Shenzhen}
  \country{China}}
\email{allengu@webank.com}

\author{Xin Yang}
\authornotemark[1]
\orcid{0000-0002-0406-6774}
\affiliation{%
  \institution{School of Computing and Artificial Intelligence, Southwestern University of Finance and Economics}
  \city{Chengdu}
  \country{China}
  \postcode{611130}
}
\email{yangxin@swufe.edu.cn}

\author{Bingjun Wei}
\orcid{0009-0008-8696-6557}
\affiliation{%
  \institution{School of Computing and Artificial Intelligence, Southwestern University of Finance and Economics}
  \city{Chengdu}
  \country{China}
  \postcode{611130}
}
\email{dutu666cai@gmail.com}

\author{Haoyang Liang}
\orcid{0009-0007-8822-8569}
\affiliation{%
  \institution{School of Computing and Artificial Intelligence, Southwestern University of Finance and Economics}
  \city{Chengdu}
  \country{China}
  \postcode{611130}
}
\email{hpsigema@gmail.com}

\author{Xiangkun Wang}
\orcid{0009-0005-5118-0122}
\affiliation{%
  \institution{School of Computing and Artificial Intelligence, Southwestern University of Finance and Economics}
  \city{Chengdu}
  \country{China}
  \postcode{611130}
}
\email{xiangkunwang18@gmail.com}

\author{Tianrui Li}
\orcid{0000-0001-7780-104X}
\affiliation{%
  \institution{School of Computing and Artificial Intelligence, Southwest Jiaotong University}
  \city{Chengdu}
  \country{China}
  \postcode{611130}
}
\email{trli@swjtu.edu.cn}

\renewcommand{\shortauthors}{Xuemei Cao, et al.}

\begin{abstract}
Continual Learning (CL) primarily aims to retain knowledge to prevent catastrophic forgetting and transfer knowledge to facilitate learning new tasks.
Unlike traditional methods, we propose a novel perspective: CL not only needs to prevent forgetting, but also requires intentional forgetting. 
This arises from existing CL methods ignoring biases in real-world data, leading the model to learn spurious correlations that transfer and amplify across tasks.
From feature extraction and prediction results, we find that data biases simultaneously reduce CL's ability to retain and transfer knowledge.
To address this, we propose ErrorEraser, a universal plugin that removes erroneous memories caused by biases in CL, enhancing performance in both new and old tasks.
ErrorEraser consists of two modules: Error Identification and Error Erasure.
The former learns the probability density distribution of task data in the feature space without prior knowledge, enabling accurate identification of potentially biased samples.
The latter ensures only erroneous knowledge is erased by shifting the decision space of representative outlier samples.
Additionally, an incremental feature distribution learning strategy is designed to reduce the resource overhead during error identification in downstream tasks.
Extensive experimental results show that ErrorEraser significantly mitigates the negative impact of data biases, achieving higher accuracy and lower forgetting rates across three types of CL methods.
The code is available at \href{https://github.com/diadai/ErrorEraser}{https://github.com/diadai/ErrorEraser}.


\end{abstract}



\maketitle

\section{Introduction}


Continual Learning (CL) or Lifelong Learning \cite{open2024cao,ke2021achieving,han2023ieta,kim2023task} has been proposed in recent years to address the challenges posed by non-stationary and dynamically changing data for neural networks. 
Its primary objectives are: (1) Preventing catastrophic forgetting: This aims to retain learned knowledge, preventing the forgetting of previously learned tasks when learning new ones \cite{van2022three}; (2) Facilitating knowledge transfer: This seeks to leverage knowledge from previous tasks to guide new tasks, thereby improving accuracy or saving learning time \cite{ke2021achieving}.
As CL methods gain widespread attention, they have evolved into three scenarios based on different learning settings \cite{van2022three}: task incremental, class incremental, and domain incremental. 
Additionally, current popular CL methods can be categorized into three types \cite{li2023cl}: memory replay-based, parameter isolation-based, and regularization-based.

\begin{figure}[!htb]
  \centering
  \includegraphics[width=0.44\textwidth]{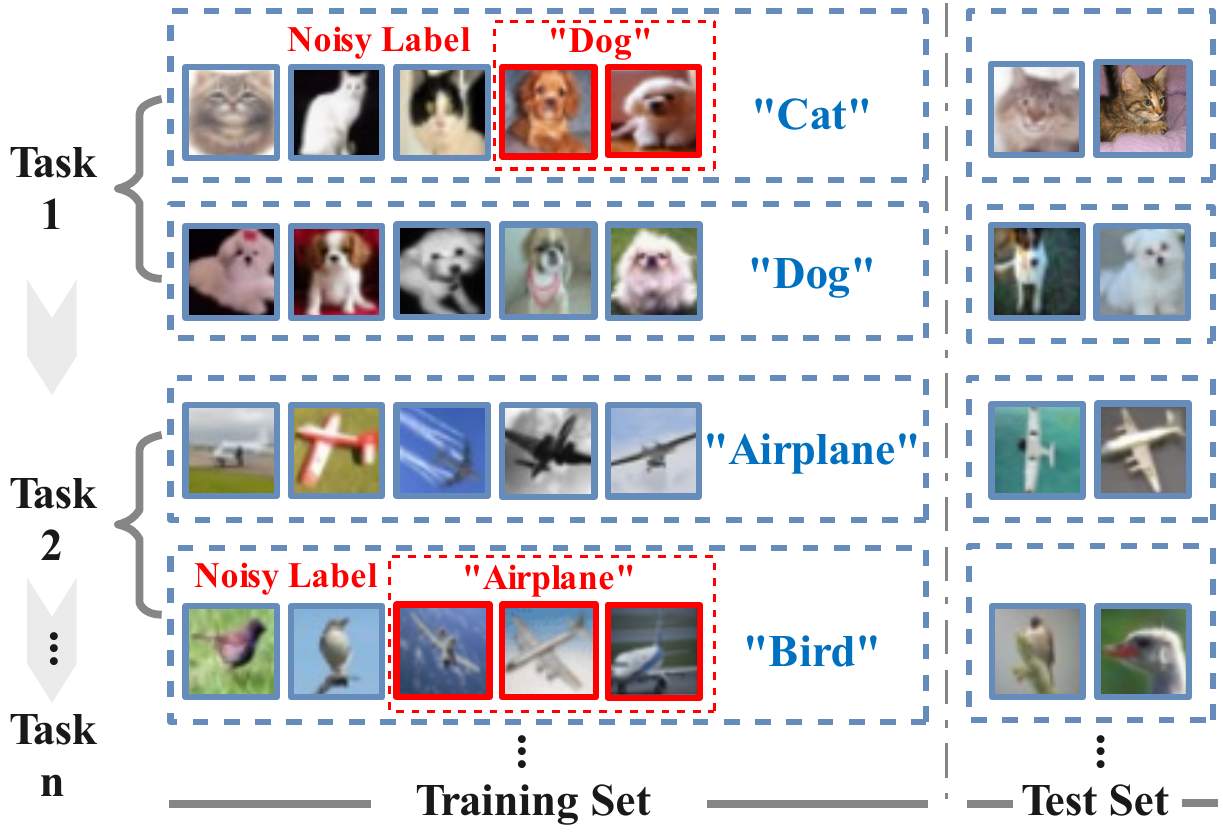}
  \caption{Diagram of the data bias.}
  \label{fig:diagram}
\end{figure}

Although current CL methods have achieved significant progress \cite{yang2024federated, lai2025order}, they heavily rely on the quality of training data \cite{kim2021continual}, often overlooking more realistic settings where data biases exist in real-world applications. During data collection, due to the large scale and the high cost of expert annotation, data often contains noisy labels, as shown in Figure \ref{fig:diagram}. This can cause the model to learn incorrect knowledge, making predictions unreliable \cite{bayasi2024biaspruner}. This presents a greater challenge for CL, as errors can accumulate and transfer across task sequences, even amplifying \cite{busch2024truth}, leading to reduced performance in both old and new tasks. This is similar to humans stubbornly retaining incorrect memories, which obstructs the acquisition of new and valuable knowledge, especially when new information conflicts with old, incorrect memories.
To address this, we provide the first in-depth experimental analysis of the specific impact of noisy labels on various CL methods (see Sect. \ref{3.1}), laying the foundation for handling data biases.



Unfortunately, existing methods cannot address the dual challenges in CL mentioned above.
A direct solution is to increase the training iterations for new tasks to continually mask the impact of noisy labels from old tasks. 
However, this approach incurs additional resource overhead and may lead to overfitting on new tasks, resulting in degraded performance on both new and old tasks (see Sect. \ref{5.3}).
Another line of solutions is to handle noisy labels such us label cleaning \cite{kim2019nlnl, yu2020unknown} and repair \cite{veit2017learning, kremer2018robust}. 
However, these methods assume the entire dataset is available for noise purification, which is infeasible in CL where historical data cannot be revisited. 
Furthermore, a few replay-based CL methods were proposed to address noisy labels, which is based on contrastive self-supervised learning \cite{kim2021continual} or pre-warmed model losses \cite{karim2022cnll}. 
Nevertheless, these approaches incur the higher training costs (see Sect. \ref{5.3}). For a detailed comparison, please refer to Sect. \ref{2.3}.

In this work, inspired by the human ability to eliminate incorrect learned knowledge through correction or forgetting \cite{ryan2022forgetting}, we propose a novel approach: achieving CL that both prevents forgetting and intentionally forgets by actively erasing already learned errors in the model.
However, combining CL with existing erasure methods presents two key challenges.
\textbf{Challenge 1:} The erroneous samples in the data stream are unknown, so before applying any forgetting method, it is essential to accurately identify these biases and their corresponding memories in the model.
\textbf{Challenge 2:} In CL, erasing the contribution of only a small number of noisy samples in the model can easily lead to a sharp decline in the performance of old tasks.
The few existing methods focus on forgetting entire classes rather than specific samples \cite{heng2024selective}, \cite{shibata2021learning}. 
Therefore, accurately forgetting only the erroneous memories in the model, without affecting the correct knowledge, is highly challenging.

These challenges motivate us to propose the \textbf{innovative ErrorEraser plugin}, which consists of two modules: Error Identification and Error Erasure. 
The ErrorEraser plugin offers \textbf{four main advantages}.
\textbf{First}, it effectively characterizes the training data distribution by learning the probability density in a compact feature space, without requiring preprocessing or prior knowledge. This accurately identifies potential biased samples based on distribution margins (addressing Challenge 1, see Sect. \ref{4.1}).
\textbf{Second}, it efficiently forgets erroneous knowledge in the model by selectively moving the decision space of only a few representative noisy label samples. The new decision space is moved away from the decision space of normal data to avoid the loss of correct knowledge (addressing Challenge 2, see Sect. \ref{4.2}).
\textbf{Third}, the incremental feature distribution learning strategy reduces resource consumption during error identification in new tasks. Additionally, selectively fine-tuning the model with representative samples is efficient and effective (see Sect. \ref{5.3}).
\textbf{Fourth}, ErrorEraser is independent of specific CL learning strategies, making it compatible with various CL methods and ensuring good generalizability (see Sect. \ref{5.2}).
Our main contributions are as follows:

(1) We are the first to deeply investigate the dual impact of noisy labels on CL. We analyze the intrinsic reasons for performance degradation caused by noisy labels from the perspectives of model feature extraction and prediction outcomes across three different types of CL methods.

(2) We are the first to enhance CL methods using intentional forgetting and propose a universal plugin, ErrorEraser. 
It effectively identifies and accurately erases potential errors already learned by the model, without the need to access historical data.

(3) We applied the ErrorEraser plugin to three types of CL methods (replay-based, regularization-based, and optimization-based). Extensive experiments on benchmark datasets validated the generality and effectiveness of ErrorEraser in forgetting erroneous knowledge caused by noisy labels.

\section{Related Work}
\subsection{Continual Learning} 
Continual learning (CL) encompasses task incremental learning \cite{aljundi2019task,shibata2021learning}, class incremental learning \cite{li2024learning,guo2023dealing}, and domain incremental learning \cite{mirza2022efficient,van2022three}. 
Existing mainstream CL methods are categorized into three types: regularization-based \cite{EWC,MAS}, replay-based \cite{shin2017continual,yoon2021online}, and model optimization-based methods \cite{LwF,ICML2022}. 
However, these methods assume that all learned knowledge in the model is beneficial, neglecting the errors introduced by data biases in real-world scenarios. These errors may also be amplified in downstream tasks of a task sequence \cite{busch2024truth}, further degrading the performance of CL.
Although recent research has begun to address this issue by introducing filtering techniques to identify clean samples within data biases \cite{kim2021continual,karim2022cnll}, allowing the model to train on correct data as much as possible, fully eliminating all problematic samples remains challenging. 
As a result, the model continues to be affected by data biases. 
Furthermore, these methods are primarily applicable to replay-based CL approaches, lacking generalizability.

To address this, this paper focuses on the correction of learned erroneous knowledge in three types of CL methods under the task-incremental learning setting (where the IDs of historical tasks are known). We propose a universal plugin method that, for the first time, tackles this issue from two perspectives: learning the correct knowledge under real-time constraints and actively erasing the erroneous knowledge that has already been learned.

\subsection{Machine Unlearning}
Machine Unlearning (MU) is an emerging data forgetting method developed in response to growing attention on model privacy protection and data privacy regulations, such as the European Union's General Data Protection Regulation (GDPR) \cite{heng2024selective}.
The primary goal of MU is to adjust a trained model to forget information learned from specified subsets of data \cite{ten2025fan}. 
Recent research has largely targeted privacy preservation in deep neural networks, categorized into class-based \cite{chen2023boundary,shibata2021learning} and instance-based data forgetting \cite{cha2024learning,gu2024unlearning}. Class-based forgetting methods, more prevalent, aim to forget all instances of a specific class while retaining the performance on other classes. Instance-based forgetting seeks to forget specific instances without affecting the performance on the remaining data, presenting a greater challenge with fewer proposed methods.

Currently, a limited body of work has integrated MU into CL to address privacy protection issues \cite{shan2024lifelong, shibata2021learning}. However, these methods assume that the data to be forgotten is known in advance, making them unsuitable for tackling the unknown data biases we aim to address, presenting an additional challenge.

\begin{table}[htb]
    \centering
    \caption{Summary of Existing Methods}
    \label{tab:Summary}
    \resizebox{0.95\linewidth}{!}{
    \begin{tabular}{c|cc|c|c}
        \toprule
        \multirow{2}{*}{\textbf{Methods}}    & \multirow{2}{*}{\textbf{CL}}  & \multirow{2}{*}{\textbf{MU}}  & \multirow{2}{*}{\textbf{Assumption}} & \textbf{Where are}   \\
        & & &  & \textbf{data bias}  \\
        \hline
                CNLL \cite{karim2022cnll} & \usym{2714}  & \usym{2718}  & -  &New task   \\
        SPR \cite{kim2021continual}  & \usym{2714}  & \usym{2718}  & -  &New task    \\
        SCL-SSR \cite{luo2023solving}  & \usym{2714}  & \usym{2718}  & -  &New task   \\
        LSF \cite{shan2024lifelong} & \usym{2714} & \usym{2714}  & Known data bias  & Old task  \\
        LSFM \cite{shibata2021learning}& \usym{2714} & \usym{2714}  & Known data bias & Old task \\
        \hline
        ErrorEraser & \usym{2714}  & \usym{2714}  & No &Old task\\
    \bottomrule
    \end{tabular}}
\end{table}

\subsection{Oversights of Existing Methods}\label{2.3}
Table \ref{tab:Summary} presents existing methods relevant to our work, with extensive experimental comparisons provided in Sect. \ref{5.3}. While LSF \cite{shan2024lifelong} and LSFM \cite{shibata2021learning} integrate continual learning (CL) and machine unlearning (MU), they assume that forgetting data are known. This assumption conflicts with the privacy constraints in CL, where the forgetting data in the old task is inaccessible due to privacy concerns. Our method, however, \textit{is capable of identifying such biases without requiring access to the original data from the old task.}
Furthermore, CNLL \cite{karim2022cnll}, SCL-SSR \cite{luo2023solving}, and SPR \cite{kim2021continual} address issues related to data biases in the new task. In this paper, \textit{we focus on the data biases present in the old task}, such as those arising from large-scale data and the high cost of expert annotation. Often, data contains noisy labels that lead to the model learning these biases. 

\section{Problem Statement}\label{3}
\subsection{Problem Phenomenon: Bias-Induced CL Degradation}\label{3.1} 

Data is the cornerstone of deep learning. When data contains noisy label biases, the model may learn incorrect knowledge, affecting its performance. 
This issue is particularly pronounced in CL. 
Specifically, we have identified \textbf{two key impacts} of noisy labels in data on CL: 
\textbf{(\romannumeral1) Knowledge Retention}: Noisy labels diminish the model's ability to retain knowledge, exacerbating catastrophic forgetting (CF) of previous tasks.
\textbf{(\romannumeral2) Knowledge Transfer}: Noisy labels cause the model to transfer erroneous knowledge, leading to degraded learning performance on new tasks.

\begin{figure}[!htb]
  \centering
  \includegraphics[width=0.47\textwidth]{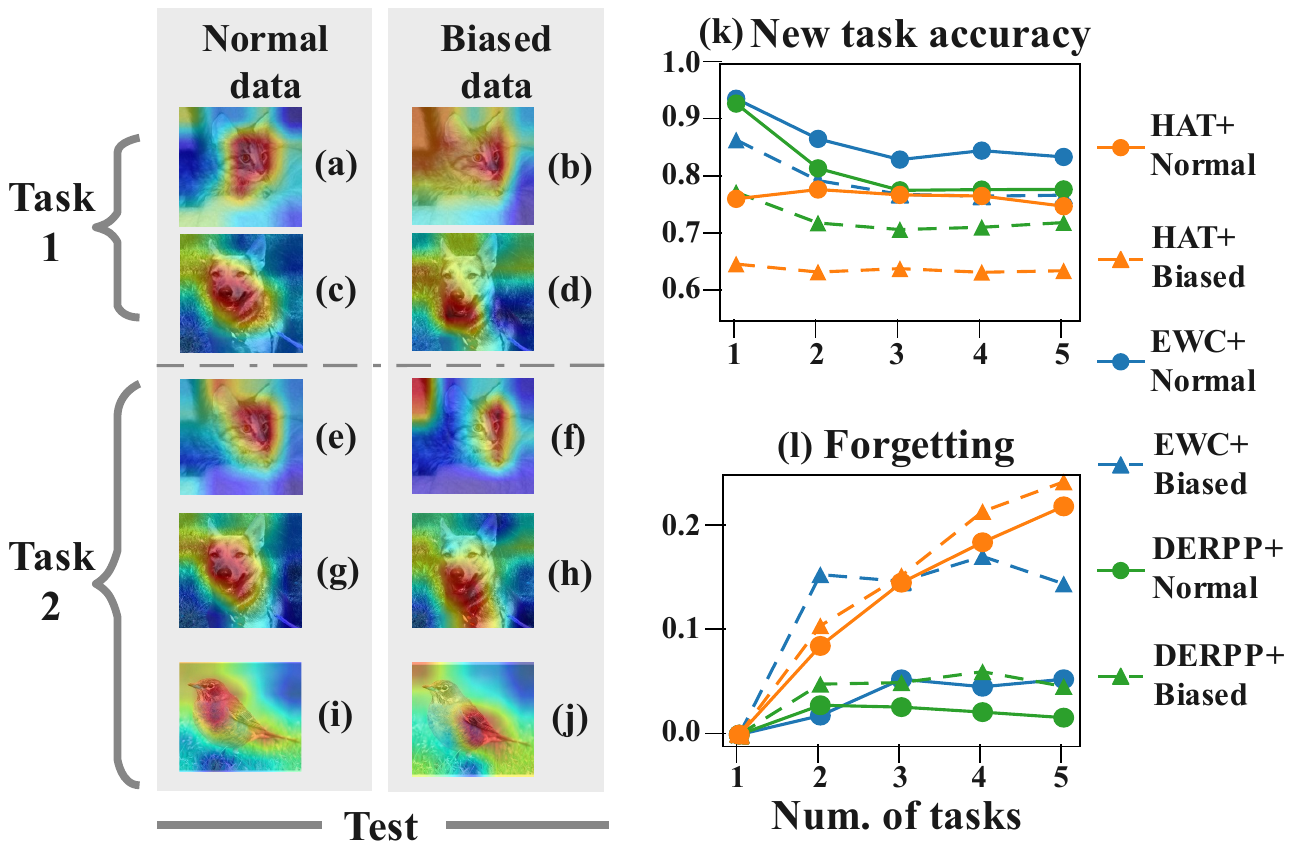}
  \caption{The impact of  noisy label biases on feature extraction (left) and prediction results (right).}
  \label{fig:impact}
\end{figure}

We conducted impact analysis experiments on CIFAR-10 with label bias and clean data, using three popular CL methods: EWC \cite{EWC}, DERPP \cite{buzzega2020dark}, and HAT \cite{serra2018overcoming}. 
Specifically, we analyzed the deep reasons behind the impact of noise on CL by visualizing model feature extraction attention and using line plots of prediction results. 
To facilitate better analysis, we set Task 1 to include noisy bias, while keeping the data in downstream tasks clean. 
The attention maps highlight the areas of focus during image prediction after the model is trained, providing an intuitive way to demonstrate whether the feature extraction was correct.
Figure \ref{fig:impact} presents the experimental results.

\textbf{Impact (\romannumeral1):}
We analyze the extent of CF in the model by comparing Task 1 test results during each task for models trained on normal data versus noisy biased data.
The feature extraction attention maps show that, after learning Task 2, noisy bias causes a significant shift in the model's attention to key features of Task 1, narrowing the focus (compare (f)(h) with (b)(d)).
This indicates significant changes in the parameter space of the feature extraction layer. 
In contrast, the model trained on normal data remains stable.
Coupled with the forgetting curve (Figure \ref{fig:impact}(l)), the model trained with noisy data has higher and steeper forgetting rates. 
Therefore, we conclude that data biases exacerbate CF and weaken CL's ability to retain knowledge in the parameter space. Works \cite{lin2023theory} and \cite{zhao2023rethinking} support the idea that changes in the feature space contribute to CF.

\textbf{Impact (\romannumeral2):}
We analyze the impact on knowledge transfer by comparing the performance of models trained on two types of data on the new task.
The attention maps show that the model trained on normal data primarily focuses on the animal's facial area (Figures (a)(c)(i)), while the model trained on noisy biased data focuses more on other body parts or the background, with more dispersed attention (Figures (b)(d)(j)). 
This suggests that noisy bias interferes with the model's ability to extract key features. 
Accuracy analysis (Figure \ref{fig:impact}(k)) further confirms this, as the model trained on noisy biased data performs significantly worse on the new task compared to the model trained on normal data. 
This indicates that erroneous knowledge introduced by noisy biases continuously transfers during CL, weakening the transfer of correct knowledge.

Therefore, erasing the erroneous knowledge learned by the model to address the dual negative impact of noisy biased data on CL is a challenging and urgent problem that needs to be solved.

\subsection{Problem Formulation}\label{3.2}


\textbf{Continual Learning:} Let $\{D_1, D_2, \dots, D_T\}$ denote a sequence of datasets corresponding to $T$ sequential tasks, where each task dataset is defined as $D_t = \{(x_t^i, y_t^i)\}_{i=1}^{n_t}$, with $x_t^i \in \mathcal{X}$ representing the input and $y_t^i \in \mathcal{Y}$ its associated class label. The goal of Continual Learning (CL) is to incrementally train a model $\mathcal{F}_{\theta_t}: \mathcal{X}_t \rightarrow \mathcal{Y}_t$ on tasks $1$ through $t$, and then update it to $\mathcal{F}_{\theta{t+1}}$ for task $t+1$ by leveraging knowledge from $\mathcal{F}_{\theta_{t}}$, such that it can correctly predict labels for any input $x$ encountered in tasks $1$ through $t+1$.


\textbf{Data Biases and Erroneous Knowledge:} Data biases may randomly emerge in tasks $1$ through $t$. Specifically, for task $t$, the dataset $D_t$ contains $k$ samples with noisy labels, denoted as $D'_t = \{x_t^i \mid 1 \leq i \leq k\}$. The true labels $\{y_t^i \mid 1 \leq i \leq k\} \in \mathcal{Y}_t$ are incorrectly annotated as $\{\hat{y}_t^i \mid 1 \leq i \leq k\}$, where $y_t^i \neq \hat{y}t^i$. When trained on such noisy data, the model $\mathcal{F}_{\theta_t}$ may capture incorrect or conflicting representations in the feature space, which we define as \emph{erroneous knowledge}, denoted by $\overline{\mathbb{K}_t}$.


\textbf{Objective:} The goal is to selectively erase the erroneous knowledge $\overline{\mathbb{K}_t}$ learned by the model $\mathcal{F}_{\theta_{t}}$ during tasks $1$ through $t$, while learning task $t+1$. Simultaneously, the model should retain and transfer correct knowledge to task $t{+}1$, thereby enhancing the training effectiveness of $\mathcal{F}_{\theta{t+1}}$.

\textbf{Constraints}: 
(1) The noisy labeled samples are randomly distributed and not known in advance.
(2) Historical data from previous tasks is inaccessible.
(3) The solution must be computationally efficient with low resource overhead.




\section{Proposed Method}\label{4}

\begin{figure*}[ht]
  \centering
  \begin{subfigure}[t]{0.49\textwidth}
    \centering
    \includegraphics[width=\textwidth]{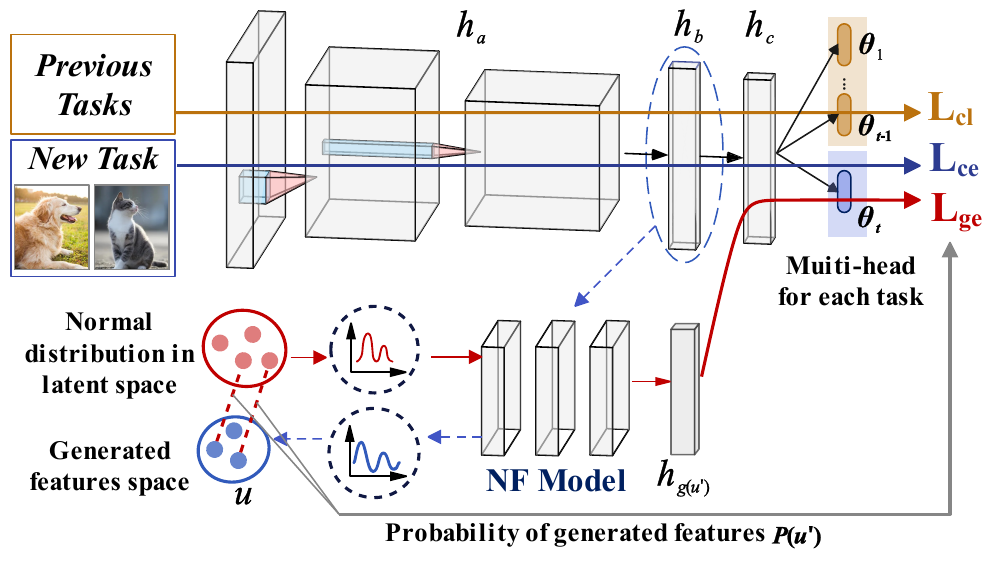}
    \caption{Error Identification.}
    \label{fig:forgetting1}
  \end{subfigure}
  \hfill
  \begin{subfigure}[t]{0.47\textwidth}
    \centering
    \includegraphics[width=\textwidth]{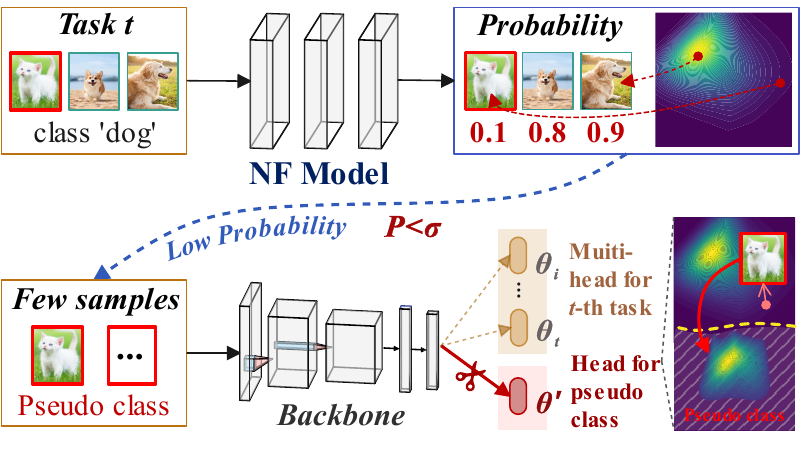}
    \caption{Error Erasure.}
    \label{fig:forgetting2}
  \end{subfigure}
  \caption{Illustrations of our ErrorEraser plugin for classifier training in CL.}
  \label{fig:combined}
\end{figure*}

In this section, we introduce ErrorEraser, a universal plugin designed to forget erroneous knowledge in CL. As shown in Figure \ref{fig:combined}, its core components are Error Recognition and Error Erasure.
Section \ref{4.1} describes the process of identifying noisy labels based on the learned data distribution. 
Section \ref{4.2} introduces how to forget erroneous knowledge in the model. 
The implementation of ErrorEraser is in Algorithm \ref{algorithm1}.

\setlength{\algomargin}{0.01\textwidth}
\begin{algorithm2e}[htb]
\caption{ErrorEraser plugin.}\label{algorithm1}
\LinesNumbered
\SetKwInOut{Input}{Input}\SetKwInOut{Output}{Output}
\Input{Task set $\{D_{train}^t, D_{valid}^t, D_{test}^t\}_{t=1}^T$, threshold $\delta$;}
\Output{Trained parameters $\theta$ and prediction model $\mathcal{F}_{\theta_{t}}$.}
\BlankLine
Initialize $\theta$;\\
\For{$task = 1$ \KwTo $T$}{
    // Error Recognition \\
    \For{each training epoch}{
        Calculate the feature space $h_b(D_{train}^t)$ of $\mathcal{F}_{\theta_{t}}$;\\
        Construct the distribution learning model $\mathcal{M}$ on $h_b(D_{train}^t)$ by Eq. \eqref{final_loss};\\
        Train $\mathcal{F}_{\theta_{t}}$ on $\{D_{train}^t, D_{valid}^t\}$ by loss $\mathcal{L}$;\\
    }
    Calculate the probability density $p$ value of $D_{train}^t$ on $\mathcal{M}$ by Eq. \eqref{probability_density};\\
    // Error Erasure \\
    Select representative sample $D_s$ according to $p$ by Eq. \eqref{Representative_Sample};\\
    \For{each forgetting epoch}
    {Fine-tune the model $\mathcal{F}_{\theta_{t}}$ on $D_s$ by Eq. \eqref{fine-tuning};\\}
}
\end{algorithm2e}

\subsection{Error Recognition}\label{4.1}
This subsection focuses on identifying potential noisy labels in the data stream to forget erroneous knowledge in the model. It begins by constructing a continual distribution learning model using an incremental feature probability density strategy to conserve training resources. A new loss function is then used to collaboratively train the backbone classification model and the distribution learning model, enabling accurate data distribution learning and reducing the impact of noisy labels in the current task. Finally, potential noisy label samples are identified based on the learned distribution.

\subsubsection{Incremental Feature Probability Density Distribution Learning}
We begin by proposing an incremental feature probability density learning strategy.
This strategy transforms complex data distributions in a compact feature space into a specified multivariate normal distribution in the latent space through a series of function transformations.
Assume the structure of the classification network model is $h = \{h_1, \ldots, h_n\}$, where $h_a = \{h_1, \ldots, h_l\}$ represents continuous feature extraction layers such as convolutional and pooling layers. $h_b = \{h_{l+1}, \ldots, h_{n-1}\}$ represents intermediate continuous feature extraction layers, such as linear layers. $h_c$ denotes the shared feature layer connected to the classification head, indicating the extraction of core features.

We first introduce the Normalizing Flows (NF) model \cite{rezende2015variational}, which can learn the probability density of data through distribution transformations.
However, directly learning the distribution on raw data poses two challenges: first, the high-dimensional nature of the data space makes the learning process time-consuming and challenging; second, the sparsity of raw data makes it difficult to capture its distribution complexity, leading to overfitting on limited training samples and poor generalization to the overall data distribution.
Therefore, we use the compact, low-dimensional feature space within the classification network model structure as input to construct the NF model. This approach not only accelerates the NF modeling process but also enables the model to better capture the distribution characteristics of the data.

Here, we use the feature space extracted by $ h_b $ as the learning object. We first transform the real feature space in $ h_b $ into a simpler prior data distribution, such as a multivariate normal distribution or a standard Gaussian distribution, through a reversible smooth mapping $ f : \mathbb{R}^d \to \mathbb{R}^d $, with the inverse mapping $ f^{-1} = g $.
Assume the prior distribution is $ u \sim p_u(u) $, and the data for the $ t $-th task is $ D_t = \{(x_t^i, y_t^i)_{i=1}^{n_t}\} $, with the corresponding $ h_b(x) \sim p_{h_b(x)}(h_b(x)) $. The distribution of the random variable $ h_b(x) $ is calculated as follows:
\begin{equation}\label{distribution_of_x}
    p_u(u) = p_{h_b(x)}(h_b(x)) \left| \det J_f(g(h_b(x))) \right|,
\end{equation}
where $ J_f $ is the Jacobian matrix of the transformation $ f $.

Then, a complex distribution can be constructed through a series of reversible mappings $ u = f_K \circ \ldots \circ f_2 \circ f_1(h_b(x)) $. At this point, using this series of reversible transformations, the probability density of $ h_b(x) $ in the prior distribution $ p_u(u) $ can be calculated as follows:
\begin{align}\label{probability_density}
\log p(x) &= \log p_u(u) + \log\left|\det\frac{\partial u}{\partial x}\right| \\
&= \log p_u(f(x)) + \sum_{l=1}^{k-1} \log\left|\det\frac{\partial f^{l+1}}{\partial f^l}\right|,\label{probability_density2}
\end{align}
where $ u = f(h_b(x)) $ is the mapping learned through the parameters $ \theta $ that transforms the distribution $ p_{h_b(x)} $. $ f^l $ denotes the input of the $ l $-th transformation in the series of mappings. 

Additionally, we use the label $ y $ as conditional information to calculate the likelihood estimate $ p_{h_b(x)}(h_b(x) | y) $. The objective of our modeling series mapping function is to maximize the likelihood of samples from the data $ D_t $. The loss function of the current task distribution learning is as follows:
\begin{align}\label{loss}
&\mathcal{L}_p(f;h_b(x)) =-\frac1n\sum_{i=1}^{n_t}\log p_{h_b(x_i)}(h_b(x_t^i) | y_t^i)\\
&=-\frac1n\sum_{i=1}^n\left(\log p_u(f(h_b(x_i))|y_i)+\sum_{l=1}^{k-1}\log\left|\det\frac{\partial f_i^{l+1}}{\partial f_i^l}\right|\right).
\end{align}

To simplify the distribution learning process in CL and avoid learning an independent distribution mapping for each task, we designed a strategy for incrementally learning the feature space distribution. This strategy uses the feature space distributions from previous tasks as constraints to guide the current task's feature space distribution learning. Thus, in CL, the improved distribution learning modeling loss based on Eq. \eqref{loss} is as follows:
\begin{align}\label{final_loss}
    \mathcal{L}_{p}(f;D_{t})=
    &-\frac{1}{|D_{t}|}\sum_{x_t^i,y_t^i\sim D_{t}}\log p_{x}(h_{b}(x_t^i)| y_t^i) \\
    &-\frac{1}{|G_{h'_b}|}\sum_{h'_{bi},y_j^i\sim G_{h'_b}}\log p_{h'_b}(h'_{bi}|y_j^i),
\end{align}
where $ y_j $ represents the class labels from tasks 1 to $ t-1 $. $ G_{h_b} $ represents the feature set of previous tasks sampled according to the condition $ y_j $ in the distribution learning model, which has the same shape as the $ h_b $ layer. This method allows for continuous learning of the feature space distributions for multiple tasks and prevents forgetting the feature distributions learned in old tasks.

\subsubsection{Distribution Learning Optimization}
Here, the focus is on obtaining high-quality feature representations from data containing noisy labels to input into the distribution learning model for a more accurate description of data distribution. We achieve this by designing a new loss function $\mathcal{L}$, which collaboratively trains the backbone classification model and the distribution learning model, allowing for real-time adjustment of input features. 
Specifically, the distribution learning model constrains the classifier's training based on the learned probability density. The updated features from the classifier are then used as input for a new round of distribution learning.

This loss function $\mathcal{L}$ comprises three components: classification loss $\mathcal{L}_{ce}$, continual learning loss $\mathcal{L}_{cl}$, and distribution learning loss $\mathcal{L}_{ge}$, as shown in Figure \ref{fig:combined}. 
The classification loss $ \mathcal{L}_{ce} $ focuses on learning new task knowledge, the continual learning loss $ \mathcal{L}_{cl} $ aims to prevent catastrophic forgetting of previously acquired knowledge, and the distribution learning loss $ \mathcal{L}_{ge} $ seeks to mitigate the impact of noisy labels. The combined loss function $ \mathcal{L} $ is defined as follows:
\begin{equation}\label{total_loss}
\mathcal{L}=\overbrace{\mathcal{L}_{ce}+\mathcal{L}_{ge}}^\text{new task}+\overbrace{\mathcal{L}_{cl}}^\text{previous tasks}.
\end{equation}

For the distribution learning loss $ \mathcal{L}_{ge} $ ($\mathcal{L}_{ce}$ and $\mathcal{L}_{cl}$, see Sect. \ref{4.2.3}), the key lies in assessing the accuracy of the feature space within the classification model.
Considering that the outlying features relative to the current task might be unreliable, we quantify the correlation between the features $ u' $ generated by the NF in the latent space and the prior distribution features $ u $. 
For ease of computation, the prior distribution in the NF adopts a multivariate normal distribution. 
Finally, we measure the correlation by calculating probabilities and design the distribution learning loss $ \mathcal{L}_{ge} $ accordingly.
We map the data $x$ to an NF latent space, resulting in $U_t = \{u_i = f(h_b(x_i))\}_{i=1}^{n_t}$, and sample a subset of generated features $U'=\{u'_i\}_{i=1}^{n'}$, then compute the correlation using the following formula:
\begin{equation}
q_{{D_{t}}}(u'_i)=\frac1{(2\pi)^{d/2}|\Sigma_{t}|^{1/2}}\exp\left(-\frac12(u'_i-\mu_{t})^T\Sigma_{t}^{-1}(u'_i-\mu_{t})\right),
\end{equation}
where $\mu_{t}=\frac{1}{|D_t|}\sum_{i=1}^{|D_t|} u_i$ represents the mean vector, and $\Sigma_{t}=\frac1{|D_t|}\sum_{i=1}^{|D_t|}\mathrm{diag}(u_i-\mu_t)\cdot\mathrm{diag}(u_i-\mu_t)$ denotes the covariance matrix.  
The $\mathrm{diag}(\cdot)$ turns the vector into a diagonal matrix.
Finally, the distribution learning loss $ \mathcal{L}_{ge} $ is defined as follows:
\begin{equation}\label{final_lge}
    \mathcal{L}_{ge}(h;U') = \frac{1}{n'}\sum_{u'\in U'} q_{D_{t}}(u') \mathcal{L}_{ce}(h_{b}(f^{-1}(u')), y_i').
\end{equation}

\subsubsection{Distribution Analysis}
Considering that abnormal features relative to the current task may be caused by noisy label samples, we evaluate the probability density value of each sample's corresponding features by Eq. \eqref{probability_density} after completing the distribution learning model. 
The likelihood of a sample being a noisy label is measured by the probability density value. Finally, samples with low probability density values at the distribution margins are detected, enabling the identification of potential noisy labels.

\subsection{Error Erasure}\label{4.2}
The error erasure of erroneous knowledge is illustrated in Figure \ref{fig:combined}(b).
The core of this process involves altering the decision space of representative noisy label samples. 

\subsubsection{Representative Sample Selection}\label{4.2.1}

After completing the t-th task, any data point $ x \in D_t $ is mapped to the NF latent space distribution through the feature extraction of network layers $ h_{a,b} $, and the corresponding probability $ p(x) $ is obtained using Eq. \eqref{probability_density} and \eqref{probability_density2}. A high probability indicates that the data is more likely to be a correct sample, while a low probability suggests the sample is less reliable and most likely a noisy labeled sample.
As shown in Figure 3(b), we visualize the distribution of the original data in the NF latent space. The central area highlighted in yellow represents the high-probability density region, consisting predominantly of correct samples. Conversely, the edges of the distribution correspond to the low-probability regions, where the sample feature space is far from the normal area, and can be considered as noisy labeled and outlier samples.

We consider the subset of samples with the lowest probability density, $ D_s $, as the representative samples for erroneous knowledge forgetting. It is defined as follows:
\begin{equation}\label{Representative_Sample}
    D_s = \{ (x_i, y_i) \in D_t \mid p(x_i) < \delta \} ,
\end{equation}
where $\delta $ is a predefined threshold that can be customized based on the data volume and other factors.
As $\delta $ increases, more samples are included in the forgetting process. 
It is important to note that the selected subset of samples should be much smaller than the current task data (i.e.,  $D_s \ll D_t $) to avoid excessive overhead.
In our implementation, $\delta$ is set based on a density $ p(x) $ percentile.

\subsubsection{Decision Boundary Shift}\label{4.2.2}

We disperse the activation of $D_s$ in the model through boundary expansion, remapping, and model pruning to forget erroneous knowledge while retaining correct knowledge. 
We intentionally assign these low-probability data $D_s$ to an additional pseudo-class in the classification model, moving its decision space to a new area distinct from the model's original decision boundaries, as illustrated in Figure \ref{fig:combined}(b). As analyzed earlier, since low-probability samples are at the margins of data distribution, altering their decision space does not affect the classification of regular samples.

Specifically, we assign a pseudo-label $\hat{y}$ to the selected $D_s$ and form new training pairs $(x_i, \hat{y}_i)$. First, we expand the model's decision space by adding a new neuron to the original classification model's output layer. 
We then fine-tune the model with these new data pairs, mapping the decision space of these likely noisy label samples to the pseudo-class.
The loss of model fine-tuning is:
\begin{equation}\label{fine-tuning}
\theta'_t=\underset{{\theta}}{\operatorname*{\operatorname*{\operatorname*{\arg\min}}}}\sum_{{(x_{i},\hat{y}_i)\in D_{s}}}\mathcal{L}_{ce}(x_{i}, \hat{y}_i, \theta_{t}),
\end{equation}
where $\theta'_t$ represents the fine-tuned model parameters.

During fine-tuning, we use the new output neuron to collect activations of noisy samples, recording and analyzing their behavior in the new neuron.
Following this, we perform pruning operations to remove the new neuron and decision space associated with the noisy samples, actively forgetting the erroneous knowledge in the model. 
After pruning, the classification model's size remains consistent with the original, maintaining the same efficiency. 
The constraint of the our entire forgetting process is:
\begin{equation}\label{constraint}
    \mathcal{F}_{\theta'_{t}}(x_i) \approx 0.
\end{equation}

Notably, the forgetting process is applied only to $D_s$ and does not allocate any samples from existing classes, so the activations of the remaining data $D_t - D_s$ do not undergo significant changes. Consequently, the model can actively forget incorrect knowledge without affecting the correct classification performance of the remaining data, thereby enhancing overall model performance.

\subsubsection{Continual Learning for New Tasks}\label{4.2.3}
After the forgetting process is completed, the learning of a new task begins. The learning loss remains as $\mathcal{L}$ in Eq. \eqref{total_loss}.
The $\mathcal{L}_{ce}$ and $\mathcal{L}_{cl}$ are not limited to specific loss functions but depend on the CL method employed.
Specifically, common $\mathcal{L}_{ce}$ in CL methods include cross-entropy loss and mean squared error loss, among others. 
The $\mathcal{L}_{cl}$ can correspond to regularization loss, replay loss, or parameter isolation loss in CL methods.
Our focus is on designing $\mathcal{L}_{ge}$ to be compatible with any $\mathcal{L}_{ce}$ and $\mathcal{L}_{cl}$, constraining the model to learn the correctly labeled samples and their probability density of the current task data as accurately as possible.
$\mathcal{L}_{ce}$ and $\mathcal{L}_{cl}$ are as follows:
\begin{align}
    &\mathcal{L}_{ce}(x;D_t) = \frac{1}{|D_t|} \sum_{i=1}^{|D_t|}\mathcal{L}_{ce}(x_i, y_i),\\
    &\mathcal{L}_{cl}(x; D_t) = \frac{1}{|D_t|} \sum_{i=1}^{|D_t|} \sum_{k=1}^{t-1} \Omega_k \mathcal{L}_{k}(x_i, y_i),
\end{align}
where $ \Omega_k $ is a weight factor that determines the importance of retaining the previous knowledge of the $ k$-th task, and $ \mathcal{L}_{k} $ represents the loss associated with the $ k $-th task. This loss can be based on various methods, such as the regularization loss in EWC: $\mathcal{L}_{k} = \frac{\lambda}{2} \sum_j (\theta_j - \theta_j^k)^2$, where $ \lambda $ is a hyperparameter controlling the strength of the regularization, $ \theta_j $ are the current parameters, and $ \theta_j^k $ are the parameters learned from the $ k $-th task.





\section{Experiments}\label{5}

\begin{table*}[t]
\caption{Results on CIFAR-10, CIFAR-100, and MNIST. Label noise is 50\%. Bold indicates that the results of applying the ErrorEraser plugin outperform the original CL method on noisy labels.} 
\label{table: Overall_performance}
\centering \arraybackslash
\resizebox{0.85\linewidth}{!}{
\begin{tabular}{c|c|cccc|cccc|ccccccc}
      &\multirow{3}*{{Methods}}      & \multicolumn{4}{c|}{CIFAR10}              & \multicolumn{4}{c|}{CIFAR100}            & \multicolumn{4}{c}{MNIST}                   \\
      &                                                    & \multicolumn{4}{c|}{\#Taks:5, \#Class:10} &\multicolumn{4}{c|}{\#Taks:5, \#Class:100} &\multicolumn{4}{c}{\#Taks:5, \#Class:10}    \\
      &                                                    &$S \uparrow$    & $A_1\uparrow$   &$A_2\uparrow$   &$F\downarrow$     & $S\uparrow$   &$A_1\uparrow$  &$A_2\uparrow$  &$F\downarrow$   &$S\uparrow$    &$A_1\uparrow$   &$A_2\uparrow$   &$F\downarrow$   \\ \hline
\multirow{6}*{\rotatebox{90}{Normal data}}                       &EWC         &57.90  &71.15    &74.88   &14.51   &26.93  &32.50  &32.97  &1.56  &81.72  &98.25   &98.63   &1.46   \\
                                                                 &MAS         &58.47  &71.56    &74.74   &12.02   &19.85  &24.33  &25.13  &4.02  &40.05  &49.42   &53.70   &13.37   \\
                                                                 &LWF         &61.12  &74.29    &77.40   &9.06    &22.43  &27.32  &28.09  &3.32  &78.96  &95.14   &98.29   &5.56   \\
                                                                 &DERPP       &63.65  &77.63    &80.17   &10.37   &16.76  &21.66  &28.07  &19.36 &81.34  &97.83   &98.78   &2.27   \\
                                                                 &HAT         &56.77  &69.62    &74.78   &14.96   &20.81  &25.40  &26.88  &4.18  &80.72  &97.09   &98.77   &3.05   \\
                                                                 &TAT         &51.89  &64.26    &70.75   &20.01   &13.56  &18.55  &29.48  &35.68 &71.22  &86.26   &88.19   &6.82     \\\hline
\multirow{6}*{\rotatebox{90}{Noisy labeled}}                     &EWC         &53.28  &65.12    &69.49   &12.07   &26.89  &32.48  &32.78  &1.56  &71.84  &87.87   &90.99   &13.65   \\
                                                                 &MAS         &54.55  &66.28    &69.65   &8.59    &19.48  &23.84  &24.83  &3.96  &69.57  &84.33   &87.95   &8.97   \\
                                                                 &LWF         &51.68  &63.16    &68.61   &13.00   &20.20  &24.77  &25.87  &4.45  &72.68  &88.58   &90.11   &9.94   \\
                                                                 &DERPP       &47.12  &58.79    &68.39   &25.9    &15.82  &20.71  &26.13  &19.72 &71.97  &87.64   &90.56   &11.00   \\
                                                                 &HAT         &45.87  &55.42    &54.55   &1.34    &12.40  &19.36  &20.75  &38.90 &56.37  &67.77   &68.29   &1.29   \\
                                                                 &TAT         &50.72  &61.98    &64.98   &10.13   &12.97  &17.64  &28.93  &34.45 &70.85  &85,72   &86.44   &4.99   \\\hline
\multirow{6}*{\rotatebox{90}{Noisy labeled}}  &\cellcolor{cyan!20}EWC+ErrorEraser   &\cellcolor{cyan!20}\textbf{64.78}  &\cellcolor{cyan!20}\textbf{78.30}   &\cellcolor{cyan!20}\textbf{78.31}   &\cellcolor{cyan!20}\textbf{3.42 }   &\cellcolor{cyan!20}\textbf{39.30}  &\cellcolor{cyan!20}\textbf{48.37}  &\cellcolor{cyan!20}\textbf{51.38}  &\cellcolor{cyan!20}10.86          &\cellcolor{cyan!20}\textbf{74.26}  &\cellcolor{cyan!20}\textbf{90.26}   &\cellcolor{cyan!20}\textbf{93.87}   &\cellcolor{cyan!20}\textbf{10.84}   \\
                                                                 &\cellcolor{cyan!20}MAS+ErrorEraser   &\cellcolor{cyan!20}\textbf{61.27}  &\cellcolor{cyan!20}\textbf{75.15}   &\cellcolor{cyan!20}\textbf{76.42}   &\cellcolor{cyan!20}11.40            &\cellcolor{cyan!20}\textbf{28.03}  &\cellcolor{cyan!20}\textbf{34.72}  &\cellcolor{cyan!20}\textbf{37.53}  &\cellcolor{cyan!20}10.06          &\cellcolor{cyan!20}\textbf{72.26}  &\cellcolor{cyan!20}\textbf{87.13}   &\cellcolor{cyan!20}\textbf{87.59}   &\cellcolor{cyan!20}\textbf{3.00 }  \\
                                                                 &\cellcolor{cyan!20}LWF+ErrorEraser   &\cellcolor{cyan!20}\textbf{58.71}  &\cellcolor{cyan!20}\textbf{71.72}   &\cellcolor{cyan!20}\textbf{72.67}   &\cellcolor{cyan!20}\textbf{8.83 }   &\cellcolor{cyan!20}\textbf{24.19}  &\cellcolor{cyan!20}\textbf{29.65}  &\cellcolor{cyan!20}\textbf{32.56}  &\cellcolor{cyan!20}7.05           &\cellcolor{cyan!20}\textbf{80.52}  &\cellcolor{cyan!20}\textbf{96.93}   &\cellcolor{cyan!20}\textbf{97.12}   &\cellcolor{cyan!20}\textbf{1.98 }  \\
                                                                 &\cellcolor{cyan!20}DERPP+ErrorEraser &\cellcolor{cyan!20}\textbf{56.62}  &\cellcolor{cyan!20}\textbf{69.66}   &\cellcolor{cyan!20}\textbf{68.57}   &\cellcolor{cyan!20}\textbf{9.26 }   &\cellcolor{cyan!20}\textbf{16.58}  &\cellcolor{cyan!20}\textbf{21.07}  &\cellcolor{cyan!20}\textbf{27.72}  &\cellcolor{cyan!20}\textbf{16.58} &\cellcolor{cyan!20}\textbf{73.97}  &\cellcolor{cyan!20}\textbf{88.91}   &\cellcolor{cyan!20}\textbf{88.29}   &\cellcolor{cyan!20}\textbf{0.29 }    \\
                                                                 &\cellcolor{cyan!20}HAT+ErrorEraser   &\cellcolor{cyan!20}\textbf{50.16}  &\cellcolor{cyan!20}\textbf{63.35}   &\cellcolor{cyan!20}\textbf{73.44}   &\cellcolor{cyan!20}33.23            &\cellcolor{cyan!20}\textbf{15.86}  &\cellcolor{cyan!20}\textbf{21.50}  &\cellcolor{cyan!20}\textbf{33.04}  &\cellcolor{cyan!20}\textbf{37.28} &\cellcolor{cyan!20}\textbf{59.35}  &\cellcolor{cyan!20}\textbf{70.58}   &\cellcolor{cyan!20}\textbf{82.41}   &\cellcolor{cyan!20}7.34   \\
                                                                 &\cellcolor{cyan!20}TAT+ErrorEraser   &\cellcolor{cyan!20}\textbf{54.52}  &\cellcolor{cyan!20}\textbf{65.86}   &\cellcolor{cyan!20}\textbf{66.43}   &\cellcolor{cyan!20}\textbf{3.16}    &\cellcolor{cyan!20}\textbf{13.82}  &\cellcolor{cyan!20}\textbf{18.81}  &\cellcolor{cyan!20}\textbf{29.01}  &\cellcolor{cyan!20}\textbf{33.58} &\cellcolor{cyan!20}\textbf{72.61}  &\cellcolor{cyan!20}\textbf{87.35}   &\cellcolor{cyan!20}\textbf{88.14}            &\cellcolor{cyan!20}\textbf{2.10 }    \\
  
\end{tabular}}
\end{table*}

\subsection{Setting}\label{5.1}

\textbf{Datasets and Implementation Details:} 
We conducted experiments on two types of noisy label data: synthetic (MNIST \cite{lecun1998gradient}, CIFAR-10 \cite{krizhevsky2009learning}, CIFAR-100 \cite{krizhevsky2009learning}) and real (WebVision \cite{song2022learning}). For the synthetic setup, following \cite{shibata2021learning}, we partitioned dataset classes into tasks to simulate CL with 5 and 10 tasks, applying asymmetric noise labeling to specific tasks \cite{li2019learning}. For WebVision with a 20\% noise rate, we divided it into 5 tasks. A Convolutional Neural Network (CNN) \cite{lecun1998gradient} was used as the classification model for all tasks, with the final layer modified to a multi-head architecture, as shown in Figure \ref{fig:combined}.

\textbf{Baselines:}
We selected three representative CL methods to validate the effectiveness of our plugin: regularization-based methods (EWC \cite{EWC} and MAS \cite{MAS}), replay-based methods (LWF \cite{LwF} and DERPP \cite{buzzega2020dark}), and optimization-based methods (TAT \cite{lin2022knowledge} and HAT \cite{serra2018overcoming}). Additionally, we compare our method with state-of-the-art data forgetting approaches LSFM \cite{shibata2021learning} and ADV+IMP \cite{cha2024learning}, as well as noisy label CL methods CNLL \cite{karim2022cnll} and SPR \cite{kim2021continual}.

\textbf{Evaluation Metric:}
In the ErrorEraser setting, our goal is to forget erroneous knowledge to enhance CL performance. Traditional metrics do not adequately evaluate performance in this context, as they overlook the combined retention and transfer capabilities of knowledge. Following \cite{shibata2021learning}, we introduce a new metric, $S$, which is the harmonic mean of three standard CL metrics: average accuracy on new tasks ($A_1$), average historical accuracy ($A_2$), and average forgetting rate ($F_k$). This metric provides a comprehensive measure of model effectiveness in a CL framework:
\begin{equation}
    S = \frac{A_1*A_2*n}{A_1+(A_2*n)+F_k},
\end{equation}
where $n$ is the total number of tasks, $A_1 = \frac{1}{n} \sum_{k=1}^n a_{k,k}, A_2 = \frac{1}{n} \sum_{k=1}^n a_{k,n}, F_k = \frac{1}{n} \sum_{k=1}^n f_k^n$, with $f_k^n = \frac{1}{(n-k)}\sum_{i=k}^n (a_{k,i}-a_{k,n})$, and $a_{k,n}$ denoting the accuracy of task $k$ after learning task $n$. 
The values of $ A_1 $, $ A_2 $, and $ F_k $ all range between [0,1].
We randomly shuffled the task order and conducted experiments five times, reporting the overall average for each metric.

\subsection{Effectiveness Guarantee}\label{5.2}

\textbf{Overall performance:} 
Table \ref{table: Overall_performance} and Table \ref{tab:webvision} present performance comparisons on synthetic and real data, respectively. Table 1 shows the results for all methods on both clean and noisy label data. It is clear that, regardless of the CL method used, including the ErrorEraser plugin significantly improves the model's performance in terms of accuracy and forgetting rate. In some cases, the results even surpass those obtained with clean data. Notably, (1) some CL methods exhibit lower forgetting on noisy data compared to clean data. This is because, as defined by the forgetting metric $F$, these methods produce similar accuracy at learning time and at final evaluation, resulting in small gaps and thus artificially low forgetting scores. (2) On certain datasets, even with our plugin applied, the HAT-based method shows slightly higher forgetting. This is because, although our plugin improves accuracy, HAT's rigid task-specific masking limits backward transfer, which can amplify forgetting.
Furthermore, the results in Table \ref{tab:webvision} further demonstrate that the ErrorEraser plugin effectively addresses the erroneous knowledge introduced by real data. This improvement is mainly due to the two modules in our plugin, which are closely aligned with the two goals of CL. The error identification module helps discard conflicting or irrelevant knowledge during the knowledge retention process, while the error erasure module further eliminates potential internal biases in the model. The appendix provides detailed experimental results for different noise label rates, task numbers, and the number of categories within tasks.
\begin{table}[htb]
    \centering
    \caption{Results on the real-world noisy dataset WebVision}
    \label{tab:webvision}
    \resizebox{0.78\linewidth}{!}{
    \begin{tabular}{c|cccc}
        \toprule
        \textbf{Methods} & \textbf{$S \uparrow$} & \textbf{$A_1 \uparrow$}  & \textbf{$A_2 \uparrow$}   & \textbf{$F \downarrow$} \\ 
        \hline
        EWC & 42.69 & 52.08 & 50.70 & 3.70 \\ 
        LWF & 41.05 & 50.21 & 48.36 & 3.69 \\ 
        HAT & 41.86 & 50.77 & 50.93 & 3.36 \\ \hline
        \cellcolor{cyan!20}EWC+ErrorEraser & \cellcolor{cyan!20}\textbf{43.72} & \cellcolor{cyan!20}\textbf{53.44} & \cellcolor{cyan!20}\textbf{51.25} & \cellcolor{cyan!20}\textbf{3.54} \\ 
        \cellcolor{cyan!20}LWF+ErrorEraser & \cellcolor{cyan!20}\textbf{42.02} & \cellcolor{cyan!20}\textbf{51.38} & \cellcolor{cyan!20}\textbf{48.70} & \cellcolor{cyan!20}\textbf{2.89} \\ 
        \cellcolor{cyan!20}HAT+ErrorEraser & \cellcolor{cyan!20}\textbf{42.75} & \cellcolor{cyan!20}\textbf{51.92} & \cellcolor{cyan!20}\textbf{51.21} & \cellcolor{cyan!20}\textbf{3.01} \\ 
        \bottomrule
    \end{tabular}}
\end{table}

\begin{figure}[h]
    \centering
    \begin{subfigure}[b]{0.14\textwidth}
        \centering
        \includegraphics[width=\textwidth]{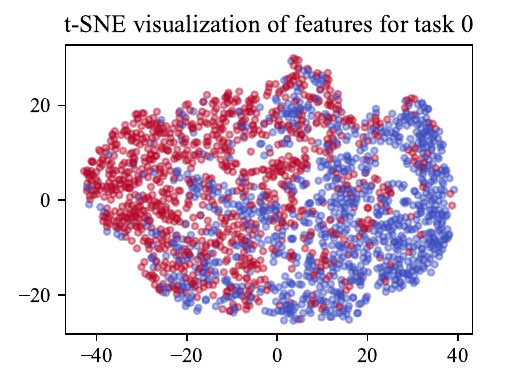}
        \caption{Task 1 + Noise}
        \label{fig:t0+mis}
    \end{subfigure}
    \hspace{0.01\textwidth}
    \begin{subfigure}[b]{0.14\textwidth}
        \centering
        \includegraphics[width=\textwidth]{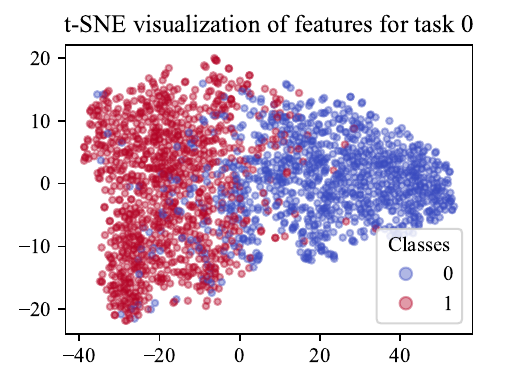}
        \caption{Task 1 $\rightarrow$ Taks 1}
        \label{fig:t0+ErrorEraser}
    \end{subfigure}
    \hspace{0.01\textwidth}
    \begin{subfigure}[b]{0.14\textwidth}
        \centering
        \includegraphics[width=\textwidth]{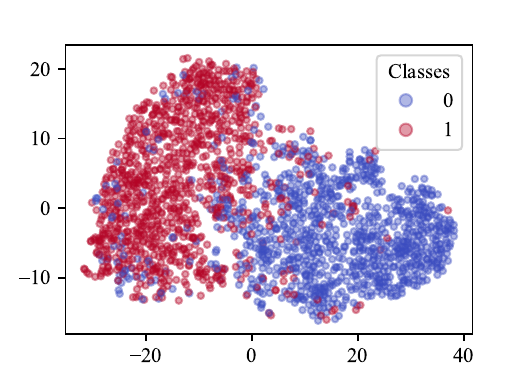}
        \caption{Task 2 $\rightarrow$ Taks 1}
        \label{fig:task2_ErrorEraser}
    \end{subfigure}


    \begin{subfigure}[b]{0.14\textwidth}
        \centering
        \includegraphics[width=\textwidth]{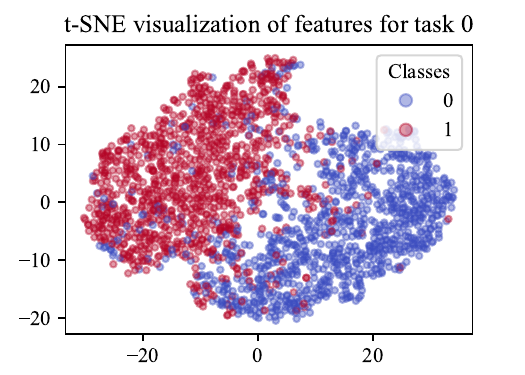}
        \caption{\small Task 3 $\rightarrow$ Taks 1}
        \label{fig:t1+ErrorEraser}
    \end{subfigure}
    \hspace{0.01\textwidth}
    \begin{subfigure}[b]{0.14\textwidth}
        \centering
        \includegraphics[width=\textwidth]{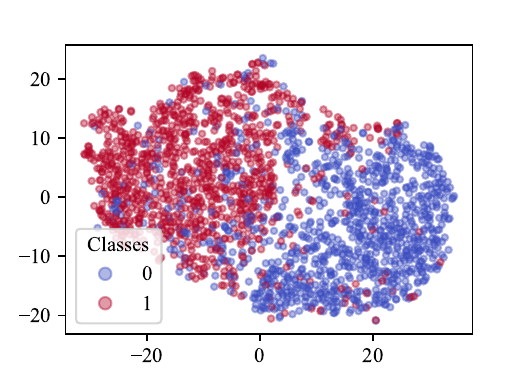}
        \caption{\small Task 4 $\rightarrow$ Taks 1}
        \label{fig:task4_ErrorEraser}
    \end{subfigure}
    \hspace{0.01\textwidth}
    \begin{subfigure}[b]{0.14\textwidth}
        \centering
        \includegraphics[width=\textwidth]{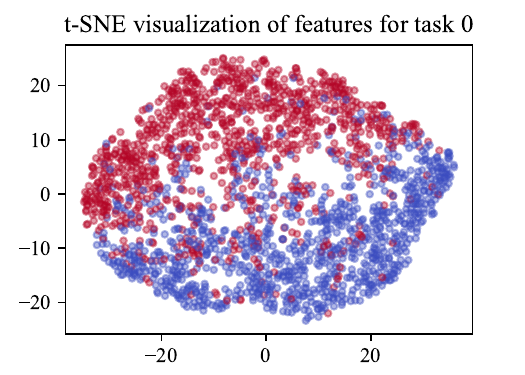}
        \caption{\small Task 5 $\rightarrow$ Taks 1}
        \label{fig:t2+ErrorEraser}
    \end{subfigure}
    \caption{t-SNE visualization of feature extraction for Task 1 (training data contains 50\% noisy labels) in the CIFAR-10 test dataset. (a) shows feature extraction for Task 1 using the original EWC model. (b)–(f) show feature extraction for Task 1 after completing Tasks 1 to 5 using ErrorEraser.}
    \label{fig:t-sne}
\end{figure}

\textbf{Visualization of Decision Space:}
To better understand how decision boundaries change with noisy label data, we visualized the decision boundaries for each task and historical task 1 after each task in Figure \ref{fig:t-sne}. With the application of the ErrorEraser plugin, we observed that the boundaries between the two classes of Task 1 data remained clear after each task, illustrating that the Error Identification module effectively prevents catastrophic forgetting. Additionally, the decision boundaries for new tasks remained compact, indicating that the Error Erasure module further eliminates erroneous knowledge, facilitating the transfer of correct knowledge to guide the learning of new tasks.

\begin{figure}[htb]
  \centering
  \includegraphics[width=0.41\textwidth]{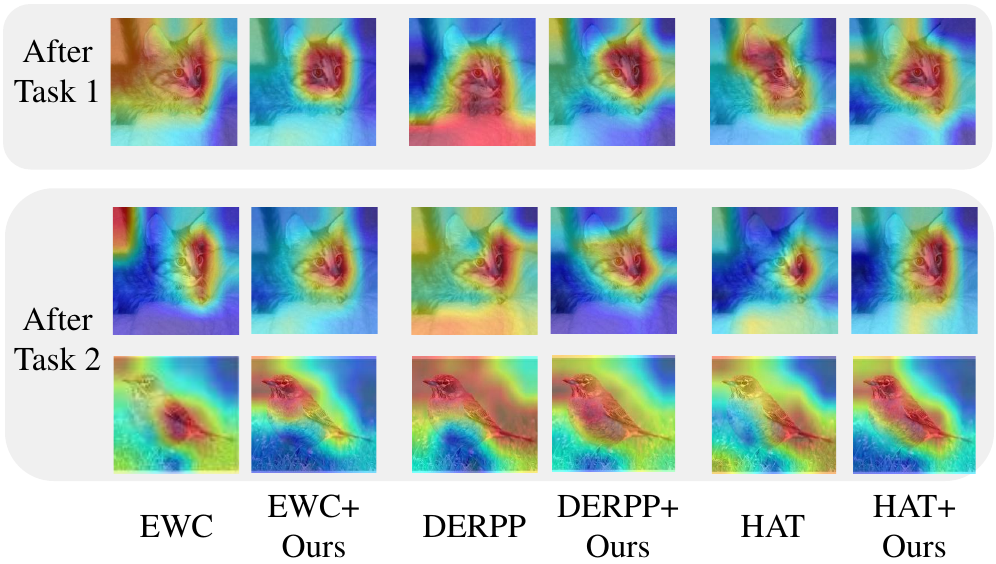}
  \caption{The attention map of three CL methods before and after applying ErrorEraser on noisy labels.}
  \label{fig:cam}
\end{figure}

\textbf{Attention Map:}
To better illustrate the impact of noisy labels, we visualized attention maps of CL models trained on the CIFAR-10 dataset with noisy labels, both with and without the ErrorEraser plugin. The highlighted regions indicate key areas for concept prediction. Each column in Figure \ref{fig:cam} represents different CL methods, and each row shows tests after completing various tasks.
We observed that after completing Task 1, methods like EWC and DERPP focused heavily on background information, with attention still misaligned from facial features. After Task 2, attention on Task 1 test data (cat) sharply decreased, focusing only on minimal facial regions. For Task 2 test data (bird), the models failed to capture comprehensive facial information. In contrast, with ErrorEraser, the models consistently focused on the core facial regions after completing both tasks. This further validates the effectiveness of the ErrorEraser plugin in erasing erroneous knowledge.

\subsection{Comparison with other methods}\label{5.3}
To evaluate the effectiveness and efficiency of our method, we also compared it with two traditional noise-handling CL methods, CNLL and SPR, which focus on constraining the model's learning of errors, while we emphasize correcting errors after the model has learned them.
In particular, we counted the running time, memory overhead, and model size of each method for training a single task.
Furthermore, we also compared other forgetting strategies.

\textbf{Effectiveness:} 
As shown in Table \ref{tab:data_forgetting_methods}, regarding data forgetting methods, LSFM targets entire classes rather than specific samples, making it less effective at selectively forgetting erroneous knowledge. The ADV+IMP method uses adversarial perturbations for sample-level forgetting, but this negatively impacts the performance of normal samples.

\begin{table}[htb]
    \centering
    \caption{Results of other data forgetting methods and noisy label CL methods.  Label noise is 50\%.}
    \label{tab:data_forgetting_methods}
    \resizebox{0.9\linewidth}{!}{
    \begin{tabular}{c|c|ccccc}
        \toprule
     \textbf{Type}     & \textbf{Methods}  & \textbf{$S \uparrow$} & \textbf{$A_1 \uparrow$}  & \textbf{$A_2 \uparrow$}   & \textbf{$F \downarrow$} \\
        \hline
        \multirow{2}{*}{{Data forgetting}} & LSFM     & 28.21   & 36.54   & 44.62    & 29.30  \\
                                           & ADV+IMP  & 39.42   & 48.95   & 59.72    & 23.21  \\
       \hline
        \multirow{2}{*}{{Noisy label}} & CNLL     & 57.66   & 71.44   & 71.95    & 14.47  \\
                                       & SPR      & 58.41   & 71.49   & 73.86    & 11.20   \\
        \hline
        \cellcolor{cyan!20}\textbf{Ours} & \cellcolor{cyan!20}\textbf{ErrorEraser} & \cellcolor{cyan!20}\textbf{64.78} & \cellcolor{cyan!20}\textbf{78.31} & \cellcolor{cyan!20}\textbf{78.30} & \cellcolor{cyan!20}\textbf{3.42} \\
        \bottomrule
    \end{tabular}}
\end{table}

Compared to CNLL and SPR, our method outperforms in preventing catastrophic forgetting and enhancing new task learning. CNLL and SPR mainly focus on mitigating catastrophic forgetting in old tasks but overlook the impact of noisy labels on new tasks. Specifically, CNLL’s approach of differentiating noisy labels based on loss from a warm-up model is insufficient. Additionally, CNLL and SPR, being replay-based methods, lack the generalizability of our approach. 
This further underscores the innovation of our method in correcting the erroneous knowledge already learned by the model.

\textbf{Efficiency:} 
The resource overhead of these methods is shown in Table \ref{tab:cost}. ``Original" represents the baseline CL method without additional measures. ``Cover" refers to the method of repeatedly training new task data (three times) to mitigate the impact of noisy labels. The results show that our method is closest to ``Original" in terms of time, memory, and model size. Specifically, ``Cover" consumes three times the resources of ``Original," proportional to the number of additional training iterations. CNLL and SPR show significant overhead in time and memory, with SPR being particularly resource-intensive due to its extensive data augmentation for contrastive self-supervised learning. Thus, our plugin is more efficient.

\begin{table}[htb]
    \centering
    \caption{Resource cost comparison}
    \label{tab:cost}
    \resizebox{0.9\linewidth}{!}{
    \begin{tabular}{c|cccccc}
        \toprule
     \textbf{Methods}    & \textbf{Time} (s)  & \textbf{Memory} (MB)  & \textbf{Model Size} (MB) \\
        \hline
        Original & $\sim 4\times10^1$ &$\sim 3\times10^2$   & 41.7  \\
        Cover & $\sim1.2 \times10^2$  &$\sim 9\times10^2$   & 125.10 \\
        CNLL  & $\sim 7\times10^2$  & $\sim 2\times10^3$  & 42.9   \\
        SPR   & $\sim 2\times10^4$  & $\sim 5\times10^3$  & 92   \\
        \cellcolor{cyan!20}\textbf{ErrorEraser} & \cellcolor{cyan!20}$\mathbf{\sim 7\times10^1} $  & \cellcolor{cyan!20}$\mathbf{\sim 4\times10^2} $  & \cellcolor{cyan!20}\textbf{51.7}   \\
    \bottomrule
    \end{tabular}}
\end{table}

\subsection{Ablation Study}\label{5.4}
We conducted an ablation experiment using EWC as the representative method under a 30\% noisy label setting, with results shown in Table \ref{tab:components}. The findings indicate that the complete ErrorEraser plugin performs best, highlighting the contribution of each component to the method's overall effectiveness. 
First, the $\mathcal{L}_{ge}$ loss in Module 1 helps CL efficiently learn the data distribution during each task and restricts training to high-probability density samples, reducing the impact of noisy labels. Module 2's active forgetting strategy further improves performance by guiding new task learning and preventing catastrophic forgetting of old tasks. This validates the effectiveness of our method in forgetting erroneous knowledge.


\begin{table}[htb]
    \centering
    \caption{Performance comparison with different component combinations. Module 1 is error identification, and Module 2 is error erasure.}
    \label{tab:components}
    \resizebox{0.95\linewidth}{!}{
    \begin{tabular}{ccccc|ccccc}
        \toprule
        & \multicolumn{4}{c}{\textbf{Components}} & \multicolumn{4}{c}{\textbf{CIFAR10}} \\
        \cmidrule(r){2-5} \cmidrule(r){6-9}
        & \multicolumn{3}{c}{\textbf{Module 1}}    & \multirow{2}*{\textbf{Module 2}}  &\multirow{2}*{\textbf{$S \uparrow$}} &\multirow{2}*{\textbf{$A_1 \uparrow$}}&\multirow{2}*{\textbf{$A_2 \uparrow$}}&\multirow{2}*{\textbf{$F \downarrow$}} \\
        & $\mathcal{L}_{ce}$ & $\mathcal{L}_{ge}$  & $\mathcal{L}_{cl}$   & &  &  &  &  \\
        \hline   
        & \checkmark        &                    &                    &            &65.04       &	86.17     &62.85        & 15.91  \\
        & \checkmark        & \checkmark         &                    &            &66.31       &  86.94      & 64.73       & 13.72  \\
        & \checkmark        & \checkmark         &                    & \checkmark &65.56       & 85.43       & 65.29        &13.48  \\
         & \checkmark        &                    & \checkmark         &           &66.14       & 81.69        & 80.00         &12.34   \\
         & \checkmark        & \checkmark         & \checkmark         &           & 68.17      & 82.67        &81.64        &4.09    \\
         & \cellcolor{cyan!20}\checkmark        & \cellcolor{cyan!20}\checkmark         & \cellcolor{cyan!20}\checkmark         &\cellcolor{cyan!20}\checkmark  & \cellcolor{cyan!20}\textbf{69.25}     & \cellcolor{cyan!20}\textbf{83.86 }    &  \cellcolor{cyan!20}\textbf{82.39}     &  \cellcolor{cyan!20}\textbf{3.01}  \\
        \bottomrule
    \end{tabular}}
\end{table}

\subsection{Sensitivity Analysis}\label{5.5}
\begin{figure}[!htb]
  \centering
  \includegraphics[width=0.45\textwidth]{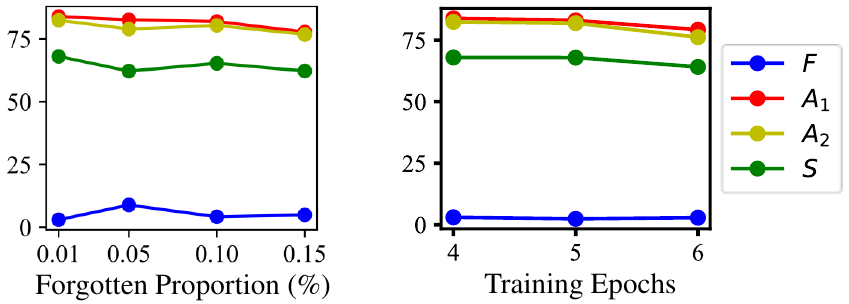}
  \caption{Performance on different parameters.}
  \label{fig: para}
\end{figure}

We analyzed the sensitivity of hyperparameters, as shown in Figure \ref{fig: para}, including the amount of data used for forgetting (i.e., the number of samples selected by Eq.\eqref{Representative_Sample}, which correspond to the lower percentile of the probability density distribution)  and the number of fine-tuning epochs. As the proportion of data used for forgetting increases, overall performance slightly decreases, as the likelihood of including normal samples increases, which may result in the erasure of correct knowledge. Similarly, increasing the number of fine-tuning epochs causes a small performance drop, as excessive epochs lead to overfitting on noisy labels. Overall, the impact of these parameters is minimal, demonstrating that the ErrorEraser plugin is robust to parameter sensitivity.




\section{Conclusion}\label{6}
This paper explores the dual impact of noisy labels on CL and introduces ErrorEraser, a universal plugin designed to address this issue. ErrorEraser learns the probability density distribution of data to effectively identify potential noisy labels in the task stream. By selectively adjusting the decision space of representative noisy samples, it accurately erases erroneous knowledge in the model, mitigating the impact of noisy labels on knowledge retention and transfer in CL. Additionally, we propose an incremental data distribution learning strategy to continuously capture each task's data distribution, optimizing resource usage. We apply ErrorEraser to three popular CL methods and conduct extensive experiments on benchmark datasets, demonstrating its effectiveness, efficiency, and generality.



\bibliographystyle{ACM-Reference-Format}
\balance
\bibliography{SFBCL}
\clearpage


\section{Experiments}

\subsection{Setting}
MNIST has 10 classes with 60,000 training images and 10,000 test images.
CIFAR-10 contains 50,000 training images and 10,000 test images overall.
CIFAR-100 comprises 100 classes of 50,000 images for training and 10,000 for testing.

\textbf{Noisy label setting:} On one hand, we introduced artificially mislabeled data in existing datasets. This refers to the asymmetric noisy label settings from \cite{kim2021continual, patrini2017making}, mapping similar classes. Specifically, for CIFAR-10, the mappings are as follows: 
\texttt{TRUCK} $\rightarrow$ \texttt{AUTOMOBILE}, 
\texttt{BIRD} $\rightarrow$ \texttt{AIRPLANE}, 
\texttt{DEER} $\rightarrow$ \texttt{HORSE}, 
and \texttt{CAT} $\rightarrow$ \texttt{DOG}. 
For MNIST, the mappings are:
$2 \rightarrow 7$, 
$3 \rightarrow 8$, 
$5 \rightarrow 6$, 
$7 \rightarrow 1$.
Following the mapping relationships for each dataset, we randomly selected a certain proportion of samples from the classes on the left side of the mappings to be relabeled as the corresponding classes on the right side. For CIFAR-10, this involved changing the labels of 50\%, 30\%, and 10\% of the samples from TRUCK, BIRD, DEER, and CAT to AUTOMOBILE, AIRPLANE, HORSE, and DOG, respectively. 
The same approach was applied to the MNIST and CIFAR-100 datasets. Compared to previous work, our noise rate settings are more challenging, as evidenced by the comparative experiments in Section 5.3 of the main text.
On the other hand, we also considered the real-world noisy label dataset webvision, which has an estimated noise rate of 20\% \cite{song2022learning}. We used the top 14 largest classes by data size, resulting in a total of 47,784 images. We curated seven tasks with randomly paired classes.

\textbf{Implementation Details:}
We train the network for 100 epochs for each task. 
For each dataset, both new and old tasks are configured with a mini-batch size of 512, applied to both the training and testing sets. 
We utilize the Adam optimizer with a weight decay of 5e-4 and a learning rate of 0.0001. 
The model parameters are initialized using the default random initialization.
\begin{table}[htb]
    \centering
    \caption{Performance Comparison on CIFAR-10 and MNIST}
    \label{tab:cifar10_mnist}
    \resizebox{0.95\linewidth}{!}{
    \begin{tabular}{lcccc|cccc}
        \toprule
        \multirow{2}{*}{\textbf{Methods}} & \multicolumn{4}{c|}{\textbf{CIFAR-10}} & \multicolumn{4}{c}{\textbf{MNIST}} \\
        & $S$ & $A_1$ & $A_2$ & $F$ & $S$ & $A_1$ & $A_2$ & $F$ \\
        \midrule
        CoFiMA \cite{marouf2024weighted} & 57.56 & 70.34 & 72.87 & 10.55 & 72.21 & 88.25 & 90.43 & 12.11 \\ 
        CoFiMA + ErrorEraser & 62.64 & 75.61 & 76.95 & 4.02 & 74.94 & 91.04 & 93.29 & 9.15 \\ 
        ICL \cite{sun2023regularizing}& 54.78 & 67.12 & 69.34 & 10.97 & 70.26 & 85.33 & 87.65 & 8.61 \\ 
        ICL+ ErrorEraser & 56.95 & 69.41 & 71.25 & 8.51 & 71.89 & 86.96 & 87.14 & 4.32 \\ 
        \bottomrule
    \end{tabular}}
\end{table}

\subsection{Comparison with other SOTA CL methods.}
The experimental results in Table \ref{tab:cifar10_mnist} demonstrate the effectiveness of ErrorEraser in improving continual learning (CL) performance across CIFAR-10 and MNIST. By integrating ErrorEraser with CoFiMA  \cite{marouf2024weighted} and ICL \cite{sun2023regularizing}), we observe consistent improvements in overall accuracy ($S$) and task-wise accuracy ($A_1, A_2$), indicating enhanced feature retention and better adaptation. Notably, ErrorEraser significantly reduces forgetting rates ($F$), with CoFiMA+ErrorEraser lowering $F$ from 10.55 to 4.02 on CIFAR-10 and from 12.11 to 9.15 on MNIST, showcasing its ability to mitigate the impact of erroneous knowledge. Similar improvements are seen with ICL, where $F$ drops from 10.97 to 8.51 and 8.61 to 4.32, respectively. These results suggest that ErrorEraser effectively refines feature representations, stabilizes model updates, and mitigates the negative effects of noise, making it a robust and generalizable plugin for improving CL performance.

\textit{It is important to note that our goal is to propose a general plugin, ErrorEraser, suitable for various CL methods, that can accurately identify and forget erroneous knowledge learned by the model, much like humans do, rather than designing a state-of-the-art (SOTA) CL method.}




\subsection{Different Noise Label Rates}
Table \ref{table: varying_mislabeled_ratio} presents the results on CIFAR-10 under different noise label rates, specifically set at 10\% and 30\%. 
The results indicate that as the noise label rate increases, the performance of standard CL methods deteriorates significantly.
In most cases, the average accuracy on new tasks ($A_1$) and historical tasks ($A_2$) shows a marked decline. 
Interestingly, the overall forgetting ($F$) appears to decrease. However, this is not due to an improvement in the model but rather because the accuracy on each task worsens, thereby reducing the computed catastrophic forgetting. Therefore, considering the comprehensive metric (S), performance degradation is positively correlated with the noise rate.

\begin{table}[htb]
\caption{Results on CIFAR-10 with varying noise label rates.} 
\label{table: varying_mislabeled_ratio}
\centering \arraybackslash
\resizebox{1\linewidth}{!}{
\begin{tabular}{c|c|cccc|cccccccccccc}                                                                      
                                                 &\multirow{3}*{{Methods}}  & \multicolumn{4}{c|}{Noise rate = 10\%}      & \multicolumn{4}{c}{Noise rate = 30\%}      \\     
                                                 &            & \multicolumn{4}{c|}{\#Taks:5, \#Class:10} &\multicolumn{4}{c}{\#Taks:5, \#Class:10}  \\     
                                                 &            &  $S$  & $A_1$ &$A_2$   &$F$   &  $S$  & $A_1$  &$A_2$   &$F$    \\ \hline
\multirow{6}*{\rotatebox{90}{Normal data}}       &EWC         &57.90  &71.15  &74.88   &14.51 &57.90  &71.15   &74.88   &14.51   \\
                                                 &LWF         &61.12  &74.29  &77.40   &9.06  &61.12  &74.29   &77.40   &9.06   \\
                                                 &HAT         &56.77  &69.62  &74.78   &14.96 &56.77  &69.62   &74.78   &14.96   \\
                                                 &MAS         &58.47  &71.56  &74.74   &12.02 &58.47  &71.56   &74.74   &12.02   \\ 
                                                 &TAT         &51.89  &64.26  &70.75   &20.01 &51.89  &64.26   &70.75   &20.01   \\
                                                 &DERPP       &63.65  &77.63  &80.17   &10.37 &63.65  &77.63   &80.17   &10.37   \\ \hline
\multirow{6}*{\rotatebox{90}{Noisy labeled }}    &EWC         &55.82  &68.84  &73.46   &17.35 &54.35  &66.97   &71.65   &16.23   \\    
                                                 &LWF         &56.89  &69.91  &74.10   &14.27 &52.24  &63.84   &68.82   &12.50   \\    
                                                 &HAT         &42.85  &51.52  &51.03   &0.03  &44.14  &53.33   &51.28   &0.01   \\    
                                                 &MAS         &56.18  &69.08  &72.76   &14.40 &53.56  &65.22   &69.30   &10.25   \\     
                                                 &TAT         &50.42  &62.42  &68.95   &19.63 &51.37  &63.67   &70.42   &20.68   \\
                                                 &DERPP       &63.68  &77.29  &79.76   &10.21 &58.08  &70.76   &73.82   &9.80  \\ \hline
\multirow{6}*{\rotatebox{90}{Noisy labeled }}    &EWC+ErrorEraser   &68.33  &82.36  &83.53   &3.39  &66.72  &80.65   &81.511  &4.41 \\     
                                                 &LWF+ErrorEraser   &58.32  &71.66  &76.07   &15.35 &57.14  &70.21   &73.64   &14.01 \\     
                                                 &HAT+ErrorEraser   &55.79  &69.92  &75.25   &25.41 &54.86  &68.84   &75.34   &27.15 \\     
                                                 &MAS+ErrorEraser   &55.83  &69.27  &75.22   &21.25 &55.91  &68.77   &74.15   &16.51 \\     
                                                 &TAT+ErrorEraser   &53.01  &64.36  &65.58   &5.80  &56.50  &68.07   &69.12   &2.66     \\    
                                                 &DERPP+ErrorEraser &61.81  &74.80  &75.86   &4.88  &57.00  &68.36   &67.69   &1.06 \\  

\end{tabular}}
\end{table}

\begin{table}[t]
\caption{Results on CIFAR-100 for varying number of tasks. }
\label{table: varying_number_T}
\centering \arraybackslash
\resizebox{1\linewidth}{!}{
\begin{tabular}{c|c|cccc|ccccccccc}
                                                 &\multirow{2}*{{Methods}}    & \multicolumn{4}{c|}{\#Taks:2, \#Class:100}  &\multicolumn{4}{c}{\#Taks:10, \#Class:100} \\
                                                 &            &  $S$  & $A_1$  &$A_2$   &$F$   &$S$    &$A_1$    &$A_2$   &$F$\\ \hline
\multirow{6}*{\rotatebox{90}{Normal data}}       &EWC         &14.65  &22.0    &21.93   &0     &37.74  &41.87    &42.48   &4.52  \\
                                                 &LWF         &12.32  &18.87   &19.275  &1.61  &30.80  &34.33    &36.16   &7.03  \\
                                                 &HAT         &12.21  &18.55   &18.78   &0.91  &31.60  &35.10    &36.73   &5.51   \\
                                                 &MAS         &11.35  &17.29   &17.55   &1.03  &28.60  &31.90    &33.84   &7.06    \\
                                                 &TAT         &16.10  &28.04   &33.42   &21.52 &13.52  &16.49    &28.62   &46.17   \\
                                                 &DERPP       &18.21  &22.03   &22.28   &1.01  &32.14  &36.16    &41.52   &15.77   \\\hline
\multirow{6}*{\rotatebox{90}{Noisy labeled}}     &EWC         &15.29  &22.95   &22.96   &0.06  &37.63  &41.73    &42.12   &4.19    \\
                                                 &LWF         &11.63  &17.84   &18.24   &1.62  &31.16  &34.64    &36.17   &5.69    \\
                                                 &HAT         &12.53  &16.3    &17.1    &9.36  &9.56   &10.53    &10.56   &0.15    \\
                                                 &MAS         &11.84  &16.87   &17.02   &0.59  &30.41  &33.85    &35.15   &5.82    \\
                                                 &TAT         &15.38  &27.04   &32.66   &22.44 &12.25  &15.02    &28.43   &49.01    \\
                                                 &DERPP       &14.02  &22.17   &22.37   &3.79  &30.99  &35.01    &39.63   &16.42    \\ \hline
\multirow{6}*{\rotatebox{90}{Noisy labeled}}&EWC+ErrorEraser  &21.87  &33.6    &34.42   &3.27  &49.43  &54.60    &54.83   &2.67     \\
                                            &LWF+ErrorEraser  &12.06  &21.02   &20.11   &8.84  &42.65  &47.50    &49.69   &8.97     \\
                                            &HAT+ErrorEraser  &11.59  &21.62   &28.57   &27.8  &18.43  &24.04    &25.98   &15.47     \\
                                            &MAS+ErrorEraser  &12.23  &20.27   &22.66   &9.55  &36.48  &40.89    &45.32   &13.82     \\
                                            &TAT+ErrorEraser  &18.04  &29.06   &31.36   &9.20  &12.23  &16.07    &29.44   &30.21     \\
                                            &DERPP+ErrorEraser&14.20  &23.10   &22.14   &4.65  &28.24  &34.07    &36.80   &3.90     \\
\end{tabular}}
\end{table}

With the addition of our ErrorEraser plugin, the impact of noisy labels is mitigated across most methods, enhancing overall performance. 
Particularly with methods like EWC, the performance with the ErrorEraser plugin exceeds the performance of EWC on clean data. This further demonstrates the superiority of our method in forgetting erroneous or irrelevant knowledge, enabling the model to learn the correct core knowledge.

\subsection{Different Number of Tasks}

Table \ref{table: varying_number_T} presents the results on CIFAR-100 with different numbers of tasks. We set the number of tasks to 2, 5, and 10, respectively. When the number of tasks is 2, each task contains 50 classes.
The results indicate that as the number of tasks decreases and the number of classes per task increases, the performance of CL methods on normal data significantly diminishes. This is because when a task contains too many classes, such as 50, the classification model struggles to accurately classify each class.
Even under these conditions, noisy labels further degrade the performance of CL methods. However, with the application of our ErrorEraser plugin, the performance of most CL methods improves across all task settings. 
This further validates the effectiveness and generality of our method.

\subsection{Per-Task Performance}  
\begin{figure}[b]
    \centering
    \begin{subfigure}{0.2\textwidth}
        \centering
        \includegraphics[width=\textwidth]{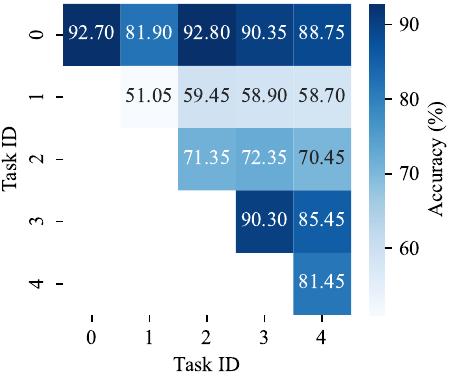}
        \caption{Accuracy on EWC}
        \label{fig:acc_ewc}
    \end{subfigure}
    \hspace{0.001\textwidth}
    \begin{subfigure}{0.2\textwidth}
        \centering
        \includegraphics[width=\textwidth]{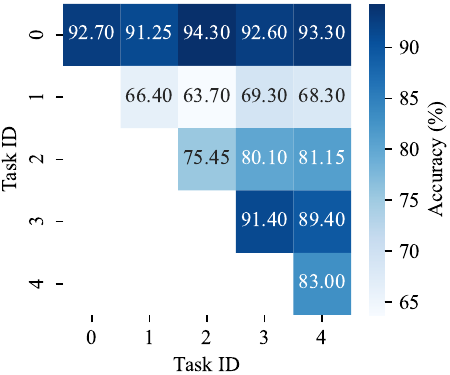}
        \caption{Accuracy on ErrorEraser}
        \label{fig:acc_ErrorEraser}
    \end{subfigure}
    \vspace{0.002\textwidth}
    \begin{subfigure}{0.2\textwidth}
        \centering
        \includegraphics[width=\textwidth]{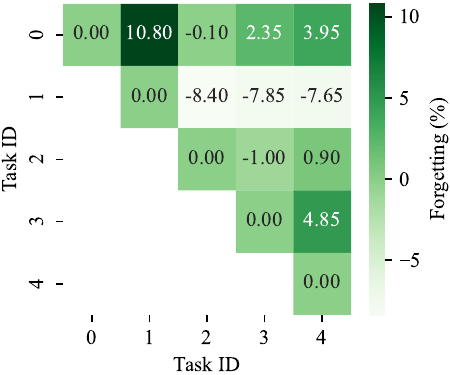}
        \caption{Forgetting on EWC}
        \label{fig:forget_ewc}
    \end{subfigure}
    \hspace{0.001\textwidth}
    \begin{subfigure}{0.2\textwidth}
        \centering
        \includegraphics[width=\textwidth]{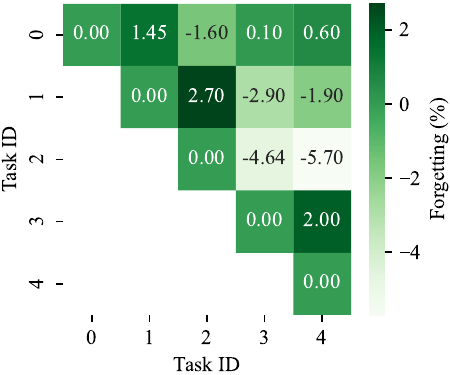}
        \caption{Forgetting on ErrorEraser}
        \label{fig:forget_ErrorEraser}
    \end{subfigure}
    \caption{Accuracy and forgetting before and after ErrorEraser.}
    \label{fig:acc_on_ewc}
\end{figure}

To illustrate the performance changes for each task, we visualize the new task accuracy and old task forgetting rates on each task after noise label introduction in Figure \ref{fig:acc_on_ewc}. 
Here, we specifically set the second task (Task ID=1) to contain 50\% noise labels to observe the results for each task.

We found that with the application of ErrorEraser, the accuracy of the second task significantly improved after its completion, and the performance of each subsequent new task also showed noticeable improvement.
Regarding forgetting, the introduction of noise labels caused the EWC method to have a forgetting rate of 10.80\% for the first task after the second task was completed. By applying ErrorEraser, the forgetting rate for the first task was reduced to 1.45\%.
This demonstrates that our method effectively forgets erroneous knowledge introduced by noisy labels, enhancing new task accuracy while preventing catastrophic forgetting of old tasks.

\subsection{Visualization of Decision Space}

\begin{figure}[htb]
    \centering
    \begin{subfigure}{0.18\textwidth}
        \centering
        \includegraphics[width=\textwidth]{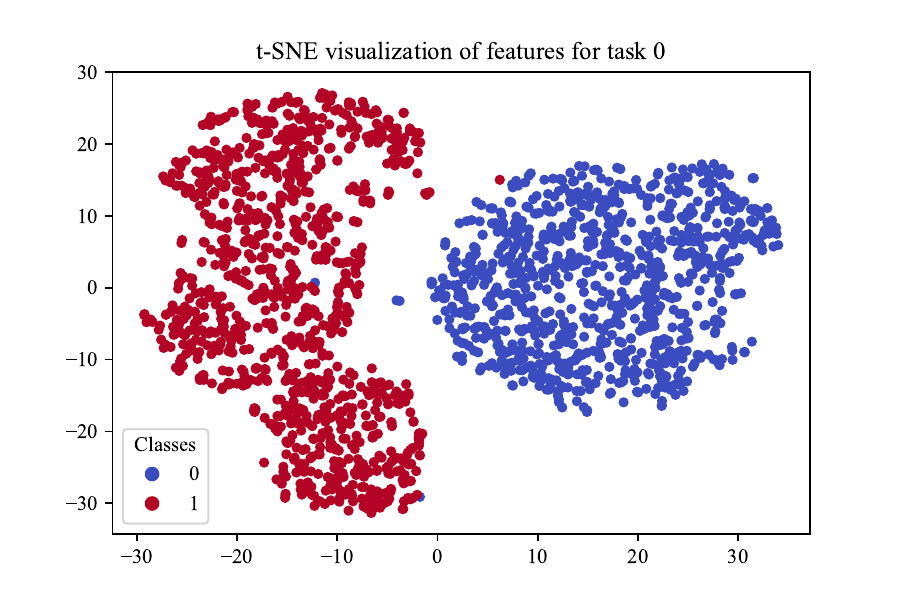}
        \caption{Task 0 + ErrorEraser}
        \label{t0}
    \end{subfigure}
    \hspace{0.02\textwidth}
    \begin{subfigure}{0.18\textwidth}
        \centering
        \includegraphics[width=\textwidth]{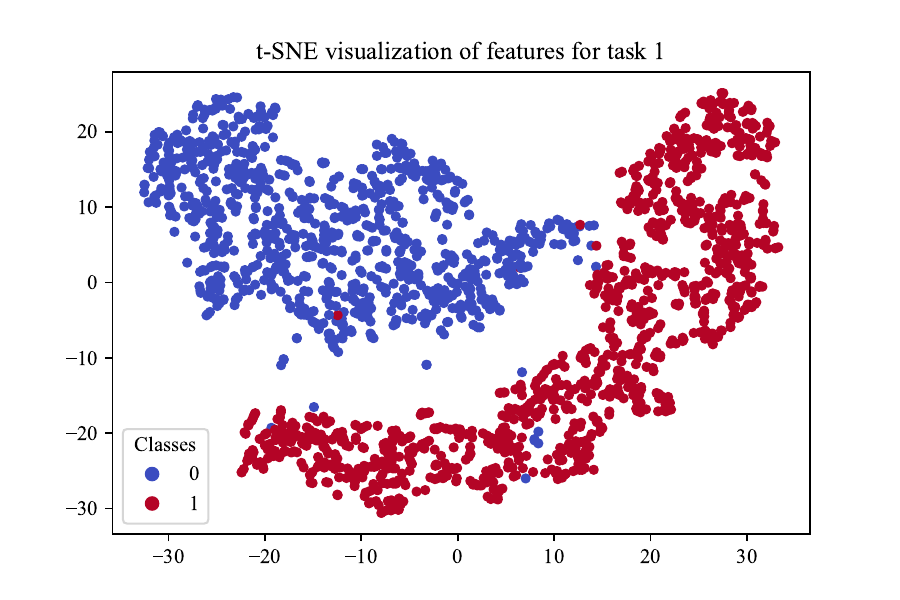}
        \caption{Task 1 + ErrorEraser}
        \label{t1}
    \end{subfigure}
    \hspace{0.02\textwidth}
    \begin{subfigure}{0.18\textwidth}
        \centering
        \includegraphics[width=\textwidth]{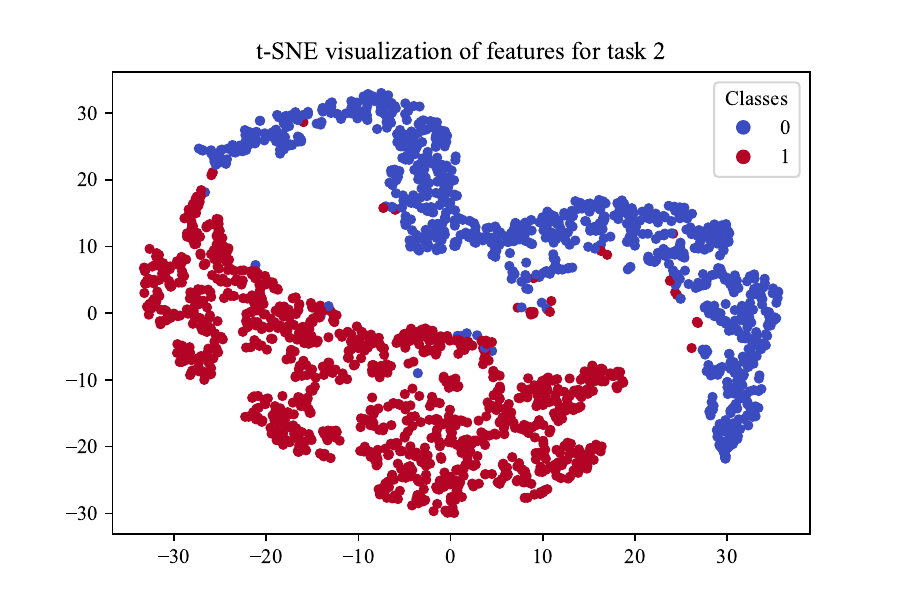}
        \caption{Task 2 + ErrorEraser}
        \label{t2}
    \end{subfigure}
    \hspace{0.02\textwidth}
    \begin{subfigure}{0.18\textwidth}
        \centering
        \includegraphics[width=\textwidth]{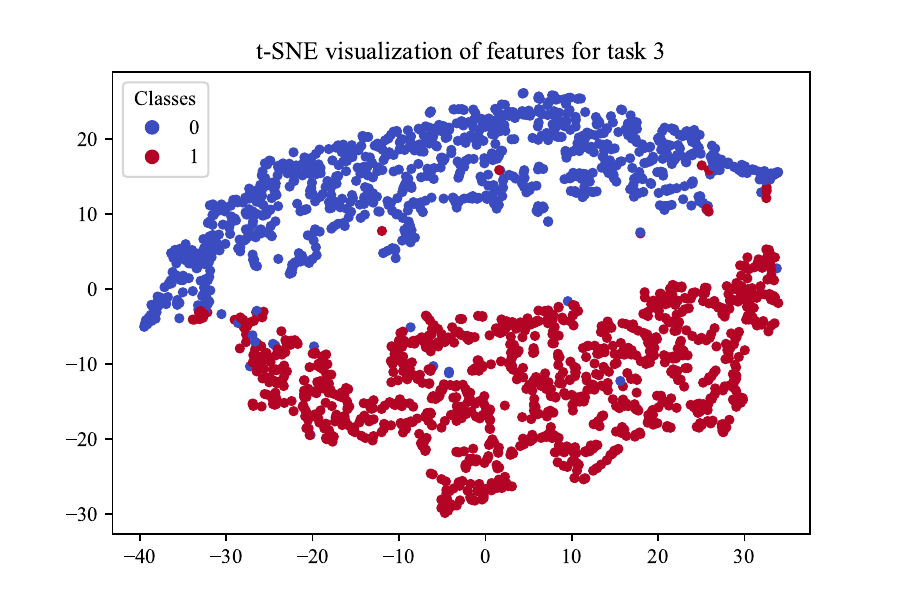}
        \caption{Task 3 + ErrorEraser}
        \label{t3}
    \end{subfigure}
    \hspace{0.02\textwidth}
    \begin{subfigure}{0.18\textwidth}
        \centering
        \includegraphics[width=\textwidth]{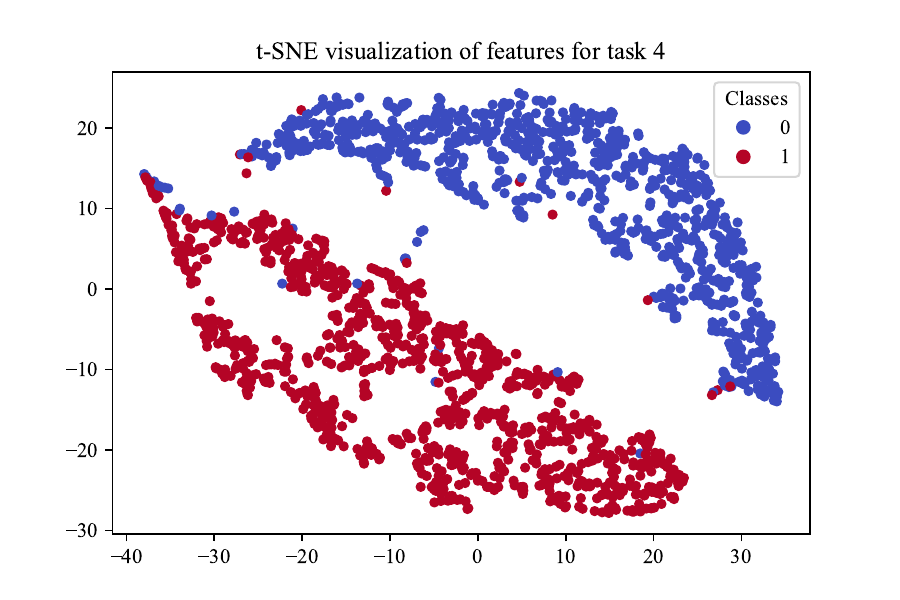}
        \caption{Task 4 + ErrorEraser}
        \label{t4}
    \end{subfigure}
\caption{t-SNE visualization of the extracted feature outputs for the each task (1-th training data contains 50\% noisy labels) in the MNIST test dataset.}
\label{fig:all_images}
\end{figure}

In Figure \ref{fig:all_images}, we present the t-SNE visualization of the feature extraction for the 1st task on the MNIST dataset after applying ErrorEraser across five tasks.
The results indicate that after each task completion, the decision boundaries between classes in the task are clear.
This further verifies that ErrorEraser effectively forgets erroneous knowledge to ensure performance on new tasks.

\section{Discussion}
The CL scenario studied in this paper is based on task incremental learning, where the task ID is known. Each task involves learning from image data, without involving multi-task learning across different data types. The classes within each task are independent and non-overlapping. Additionally, this study focuses solely on the data bias introduced by noisy labels.

In the future, we plan to explore the impact of noisy labels on CL across all three scenarios, aiming to develop more comprehensive and general solutions. We will also investigate the challenge of overlapping classes between tasks, where the impact of noisy labels may be more severe and complex. Finally, we intend to address more realistic data bias situations, such as entire classes being incorrectly labeled, noisy data (e.g., blurred or damaged images), to develop robust CL methods. We aim to continuously contribute to the continual learning community.

\clearpage
\end{document}